# Review of coreference resolution in English and Persian

Hassan Haji Mohammadi [a,1], Alireza Talebpour [b], Ahmad Mahmoudi Aznaveh [b], Samaneh Yazdani [a]

[a] Islamic Azad University Tehran North Branch, Tehran, Iran

[b] Shahid Beheshti University, Tehran, Iran

## Abstract

Coreference resolution (CR), identifying expressions referring to the same real-world entity, is a fundamental challenge in natural language processing (NLP). This paper explores the latest advancements in CR, spanning coreference and anaphora resolution. We critically analyze the diverse corpora that have fueled CR research, highlighting their strengths, limitations, and suitability for various tasks. We examine the spectrum of evaluation metrics used to assess CR systems, emphasizing their advantages, disadvantages, and the need for more nuanced, task-specific metrics. Tracing the evolution of CR algorithms, we provide a detailed overview of methodologies, from rule-based approaches to cutting-edge deep learning architectures. We delve into mention-pair, entity-based, cluster-ranking, sequence-to-sequence, and graph neural network models, elucidating their theoretical foundations and performance on benchmark datasets. Recognizing the unique challenges of Persian CR, we dedicate a focused analysis to this under-resourced language. We examine existing Persian CR systems and highlight the emergence of end-to-end neural models leveraging pre-trained language models like ParsBERT. This review is an essential resource for researchers and practitioners, offering a comprehensive overview of the current state-of-the-art in CR, identifying key challenges, and charting a course for future research in this rapidly evolving field.

## 1. Introduction

Natural language processing (NLP) is a multifaceted research field that explores the intricate relationship between computers and human language. NLP aims to develop algorithms and models that enable computers to effectively process and understand vast quantities of natural language data. This encompasses various subfields, including discourse analysis, syntactic and semantic parsing, and context understanding. The recent surge in Large Language Models (LLMs) has dramatically advanced NLP capabilities, offering potential solutions to various challenges within

---

1 Email addresses : *h.hajimohammadi@iau-tnb.ac.ir* (Hassan Haji Mohammadi), *Talebpour@sbu.ac.ir* (Alireza Talebpour) , *a_mahmoudi@sbu.ac.ir* (Ahmad Mahmoudi Aznaveh), *samaneh.yazdani@gmail.com* (Samaneh Yazdani) the field (Zhao, Zhou et al. 2023). However, NLP faces numerous challenges, one of the most significant being coreference resolution.

The discourse model consists of a collection of sentences that make sense only when combined. In computational linguistics, anaphora resolution involves finding an antecedent for pronouns and definite expressions whose interpretation depends on a previous contextual expression. Coreference resolution goes one step further and identifies all in-text expressions that refer to the same real-world entity.

Coreference resolution (CR) is a critical task in NLP, defined as "the problem of identifying all noun phrases or in-text mentions that refer to the same real-world entity" (Finkel and Manning 2008, Ng 2009, Stoyanov, Gilbert et al. 2009). Alternatively, it can be framed as "grouping all the mentions within a document into equivalent classes where all the mentions within the class refer to the same discourse entity" (Bengtson and Roth 2008, Denis and Baldridge 2008). To illustrate, consider the phrase, "The Iranian president attended the meeting of the Persian Gulf Council. He met other members after breakfast. Hassan Rouhani then attended the summit." The mentions "The Iranian president," "He," and "Hassan Rouhani" all refer to the same individual, forming a coreference chain.

*("The Iranian president" attended the meeting of the Persian Gulf Council. "He" met other members after breakfast. "Hassan Rouhani" then attended the summit.)*

CR is a binary classification problem: two mentions in the text are either coreferent or not. Despite its challenges, CR has far-reaching applications in various NLP tasks, including information extraction (McCarthy and Lehnert 1995, Zheng and Tuan 2023), text summarization (Azzam, Humphreys et al. 1999, Steinberger, Poesio et al. 2007, Antunes, Lins et al. 2018), question answering (Morton 2000), machine translation (Mitkov 1995, Vu, Kamigaito et al. 2024), sentiment analysis (Nicolov, Salvetti et al. 2008, Cambria 2016), and machine reading (Poon, Christensen et al. 2010, Bi, Liu et al. 2023).

The evolution of CR methods has significantly shifted from early rule-based systems relying on hand-crafted features (Hobbs 1978, Carbonell and Brown 1988, Rich and LuperFoy 1988) to sophisticated deep neural network models (Xia, Sedoc et al. 2020, Xu and Choi 2020, Bai, Zhang et al. 2021, Zhang, Wiseman et al. 2023, Hou, Wang et al. 2024). This review paper will examine this trajectory, comparing and contrasting these two paradigms. Several rule-based and machine-learning algorithms have been proposed in recent years. In addition, various researchers have developed different types of models, such as mention-pair (Soon, Ng et al. 2001, Sahlani, Hourali et al. 2020), cluster-based (Culotta, Wick et al. 2007), and ranking models (Denis and Baldridge 2008, Lee, He et al. 2017, Lee, He et al. 2018).

Several comprehensive reviews have summarized the progress in CR, covering algorithms, corpora, tools, and evaluation metrics (Ng 2003, Ng 2010, Ferreira Cruz, Rocha et al. 2020, Lata, Singh et al. 2020, Sukthanker, Poria et al. 2020, Stylianou and Vlahavas 2021, Lata, Singh et al.

2022, Liu, Mao et al. 2023). (Mitkov 2014) focused on anaphora resolution, a closely related subtask of CR. (Ng 2010) provided a 15-year retrospective on CR research, while (Sukthanker, Poria et al. 2020) delved into machine learning and deep learning approaches. [25] examined corpora and evaluation metrics alongside end-to-end deep learning systems. (Stylianou and Vlahavas 2021) focused on neural network-based CR and pronoun resolution. (Lata, Singh et al. 2022) provided a detailed analysis of mention detection methods. More recently, (Liu, Mao et al. 2023) presented a brief yet comprehensive survey of recent advances in CR, including analyzing evaluation metrics, datasets, and emerging technical trends.

The research direction in information extraction has recently changed from entity-based to event-based coreference resolution (De Langhe, De Clercq et al. 2022). The challenges of event coreference resolution are more significant than those of entity coreference resolution. Several neural network systems have recently been proposed in the neural event coreference resolution field (Hürriyetoğlu, Mutlu et al. 2022, Lu, Lin et al. 2022, Yu, Yin et al. 2022, Gao, Li et al. 2024). In this paper, event coreference is not investigated.

This review paper aims to contribute to the field of CR in the following ways:

- **Comprehensive Survey of CR Models:** We conduct an extensive review of 178 articles, spanning the entire spectrum of CR models, from foundational rule-based approaches to cutting-edge deep learning architectures. This survey provides a view of the CR landscape, highlighting the evolution of methodologies, key milestones, and emerging trends.
- **In-Depth Corpus Analysis:** We go beyond simply listing datasets by critically analyzing various English and non-English corpora's strengths, weaknesses, and suitability for different CR tasks. Our analysis emphasizes the need for more diverse and representative corpora, particularly for under-resourced languages like Persian. We highlight recent advancements in corpus development, such as the Mehr corpus for Persian, which addresses previous limitations and offers improved annotations.
- **Comprehensive Evaluation Metric Examination:** We examine the essential evaluation metrics used in CR research, providing clear explanations and examples. We analyze the strengths and limitations of each metric, discussing how they capture different aspects of CR performance and guide model development. We emphasize the need for more nuanced and task-specific metrics that better reflect real-world CR applications.
- **In-Depth System Review with Comparative Analysis:** We conduct a comprehensive review of CR systems, ranging from rule-based to deep learning methods, analyzing their implementation, results, strengths, and weaknesses. Our review goes beyond summarizing individual systems by providing a comparative analysis across different architectures and languages. We highlight the evolution of CR systems from hand-crafted features and pipeline approaches to end-to-end neural models that jointly learn mention detection and coreference linking. We also discuss the specific challenges and opportunities in developing CR systems for Persian.

- **Challenge Exploration and Future Directions:** We identify and discuss the key challenges hindering CR performance, such as handling ambiguous references, resolving coreferences across long distances, adapting to different domains and languages, and addressing the limitations of existing evaluation metrics. We propose future research directions, including:
    - **Exploring Alternative Architectures:** We are investigating novel architectures like graph neural networks (GNNs) for capturing complex dependencies and global context and exploring the potential of sequence-to-sequence (Seq2Seq) models for their flexibility and adaptability to different languages.
    - **Incorporating External Knowledge:** Leveraging knowledge graphs, ontologies, and domain-specific embeddings to enhance the model's understanding of entities and their relationships.
    - **Developing Explainable Models:** Creating CR models that provide transparent explanations for their decisions, building trust and enabling better analysis of their behavior.
    - **Improving Evaluation:** Developing more comprehensive and task-specific evaluation metrics that better reflect the real-world impact of CR errors.
- **Focus on Persian CR:** We dedicate a portion of this review to examining Persian CR's unique challenges and opportunities. We discuss the limitations of existing Persian systems, the need for more diverse and representative corpora, and the potential of advanced NLP techniques like pre-trained language models (e.g., ParsBERT) (Farahani, Gharachorloo et al. 2021), attention mechanisms, and GNNs for improving Persian CR. We highlight the recent development of the first end-to-end neural system for Persian pronoun resolution (Mohammadi, Talebpour et al. 2024), demonstrating these techniques' potential in addressing Persian-specific linguistic challenges.

Coreference resolution systems typically share a typical architecture, although the specific components can vary depending on the model. End-to-end deep learning systems, for instance, bypass traditional preprocessing and feature engineering steps, directly learning representations from raw text data. Figure 1 illustrates a general CR architecture, starting with data preparation and feature extraction, followed by feature selection, coreference classification, and finally, the formation of coreference chains.

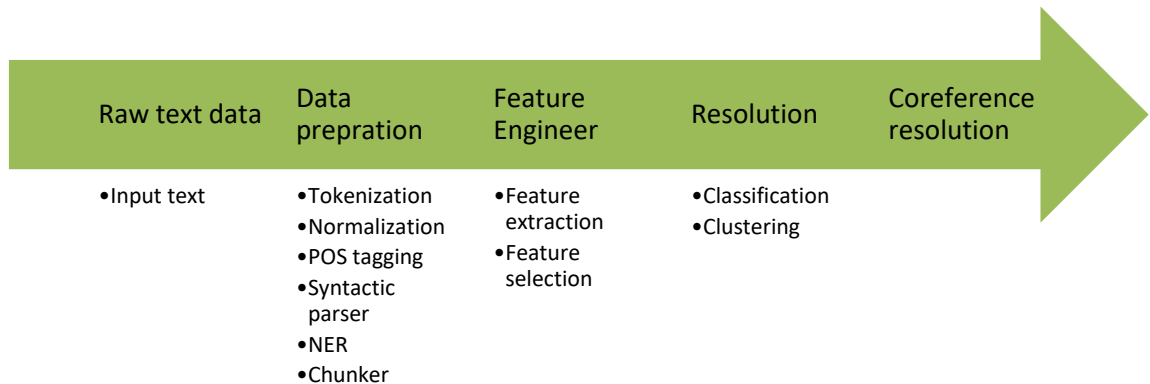

**Figure 1.** Coreference resolution architecture

The remainder of this article is structured as follows:

The remainder of this review paper is organized as follows: Section 2 provides a critical analysis of popular coreference corpora, examining their strengths, limitations, and suitability for various tasks. Section 3 delves into existing evaluation metrics for coreference resolution, including their underlying principles and practical examples. Section 4 explores the evolution of coreference algorithms from rule-based systems to cutting-edge deep learning models. Section 5 focuses on the unique challenges and opportunities in Persian coreference resolution. Section 6 synthesizes critical challenges in the field and proposes future research directions, and Section 7 concludes with a summary of key findings and insights.

## 2. Coreference corpora: A Critical Review and Analysis

Robust coreference resolution (CR) research relies on well-annotated corpora, particularly in practical applications. Corpora are indispensable for training machine learning models, evaluating system performance on large-scale data, and investigating linguistic phenomena and intertextual relationships. The availability of a suitable corpus is crucial for comparing the performance of various supervised systems in natural language processing (NLP) algorithms. Over the years, numerous coreference corpora of varying sizes and languages have been developed. These corpora differ based on their domain, annotation scheme, types of annotated relationships, size, and the specific language under study.

This section examines prominent English coreference corpora and explores corpora from other languages. However, this review does not examine task-specific corpora, such as those focusing on medicine coreference.

### 2.1. Key English Corpora: Evolution and Limitations

OntoNotes 5.0 (Hovy, Marcus et al. 2006) is English's most extensive coreference corpus, setting a standard for other languages to emulate. For instance, the RCDAT, a one million-word Persian corpus, has been developed with similar ambitions (Rahimi and HosseinNejad 2020). The MUC corpus, introduced at the Sixth and Seventh Message Understanding Conferences, was pivotal in early CR research. MUC-6 (Grishman and Sundheim 1996) and MUC-7 (Hirschman 1997) served as benchmarks for several years but were eventually superseded by larger-scale corpora. Consisting of 318 Wall Street Journal articles in SGML format, these corpora were limited in size and scope, annotating only identity relations and omitting singletons.

The ACE corpus (Doddington, Mitchell et al. 2004), developed in four versions between 2002 and 2008, expanded the range of content types and languages (English et al.). However, it faced limitations by considering only seven types of entities and needing a clear separation of training and test data, making system comparisons challenging. The MATE project (Poesio 2004) revolutionized annotation practices with its in-depth examination of annotation formulation. This framework influenced numerous subsequent corpora in English (Poesio, Delmonte et al. 2004, Hovy, Marcus et al. 2006, Pradhan, Hovy et al. 2007, Pradhan, Ramshaw et al. 2007, Poesio and Artstein 2008) and other languages (Hinrichs, Kübler et al. 2005, Hendrickx, Bouma et al. 2008, Taulé, Martí et al. 2008, Rodrıguez, Delogu et al. 2010), offering a broader range of nominal groups and relationships. The MMAX2 annotation tool (Müller and Strube 2006) facilitated this annotation process.

The CoNLL-2011 corpus (Pradhan, Ramshaw et al. 2011), based on OntoNotes 2.0, and the multilingual CoNLL-2012 corpus (Pradhan, Moschitti et al. 2012), based on OntoNotes 5.0, have become standard benchmarks for CR systems. Despite their significant size increase, these corpora have limitations, including omitting singletons and certain non-referential expressions. The GNOME corpus (Poesio 2000, Poesio 2004) was a pioneering cross-domain corpus utilizing the MATE annotation method. It was instrumental in studying centering theory and evaluating cross-domain CR algorithms. The ARRAU corpus (Poesio and Artstein 2008) furthered this work by tackling complex relationships like multi-reference resolution. This corpus, which comprises different categories, was soon adopted by the Italian corpus *LiveMemories* (Rodrıguez, Delogu et al. 2010).

**2.2. Specialized and Multilingual Corpora**

Beyond general-domain corpora, several specialized corpora have been developed for specific tasks and languages. The NP4E corpus (Hasler and Orasan 2009) and the ECB+ corpus (Cybulska and Vossen 2014) target event coreference resolution, while the ParCor and ParCorFull corpora (Guillou, Hardmeier et al. 2014, Lapshinova-Koltunski, Hardmeier et al. 2018) address parallel pronoun coreference for machine translation. These two corpora are labeled in German and English for machine translation purposes and are annotated in MMAX2 format. Moreover, the CIC corpus (Chen and Choi 2016) was developed to improve coreference in Chatbot in CoNLL format. GUM (Zeldes 2017), WikiCoref (Ghaddar and Langlais 2016), KnowRef (Emami,

Trichelair et al. 2018), PreCo (Chen, Fan et al. 2018), and LitBank (Bamman, Lewke et al. 2019) are examples of corpora for specific coreference domains. However, most of these corpora are small or were created to address only a portion of the coreference task. Consequently, they were unable to replace CoNLL-2012 as a benchmark.

The creation of non-English corpora has been attempted, except for the non-English components of ACE and CoNLL. In SEMEVAL evaluation (Ruppenhofer, Sporleder et al. 2010), the competition dataset includes a collection of Italian LiveMemories, Spanish and Catalan ANCORA (Recasens and Martí 2010), German Tüba/DZ (Hinrichs, Kübler et al. 2005), and English OntoNotes corpus. They were all converted to a standard format and given the same annotation. The University of Barcelona's ANCORA corpus (Taulé, Martí et al. 2008, Recasens and Martí 2010) results from years of effort and annotation in Spanish and Catalan in multiple steps. Its annotation standard is formatted in MATE. The RCDAT corpus (Rahimi and HosseinNejad 2020) in Persian has been annotated with approximately one million tokens using the CoNLL standard. Table 1 lists several significant coreference corpora in various languages.

## 2.3. Recent Developments

Recent research has emphasized the need for more extensive and diverse corpora that better reflect real-world language use and address biases in existing datasets. The GAP Coreference Corpus (Webster, Recasens et al. 2018) has been developed to mitigate gender bias and social stereotypes. Additionally, ongoing efforts aim to create corpora for low-resource languages and specific domains, such as the recent introduction of the Mehr corpus for Persian coreference resolution (Haji Mohammadi, Talebpour et al. 2023). Recent advancements in multilingual coreference resolution have also led to the development of corpora such as ThaiCoref (Trakuekul, Leong et al. 2024) for Thai and a Lithuanian Coreference Corpus (Žitkus 2024). Moreover, research has explored enhancing Turkish coreference resolution through insights from deep learning, dropped pronouns, and multilingual transfer learning (Arslan and Eryiğit 2024).

By offering a nuanced and critical perspective on the existing corpus landscape, this review aims to empower researchers and practitioners to make informed decisions about dataset selection, inspire the development of new corpora that address current challenges, and ultimately advance the field of coreference resolution.

**Table 1.** An overview of unrestricted coreference corpora in various languages

| Language | Corpus Name | Reference | Size |
|---|---|---|---|
| English | GNOME | (Poesio 2004) | 40k |
| | ACE-2 | - | 180k |
| | ACE-2005 | (Doddington, Mitchell et al. 2004) | 400k |
| | ACE-2007 | - | 300k |
| | ARRAU 2.0 | (Poesio and Artstein 2008) | 300k |
| | ONTONOTES 5.0 | (Pradhan, Hovy et al. 2007) | 1450k |
| | CoNLL 2012 | (Pradhan, Moschitti et al. 2012) | 1450k |
| | MUC-6 | (Grishman and Sundheim 1996) | 30k |
| | MUC-7 | (Hirschman 1997) | 25k |
| **Italian** | LIVE MEMORIES | (Rodrıguez, Delogu et al. 2010) | 600K |
| | I-CAB | (Magnini, Pianta et al. 2006) | 250K |
| | Venex | (Poesio, Delmonte et al. 2004) | 40k |
| **Spanish** | ANCORA-CO-ES | (Recasens and Martí 2010) | 400k |
| | ACE-2007 | - | 200k |
| **Hindi-Bengali** | ICON | (Sobha, Bandyopadhyay et al. 2011) | - |
| **German** | Tüba/DZ | (Hinrichs, Kübler et al. 2005) | 600k |
| **Dutch** | CUREA | (Hendrickx, Bouma et al. 2008) | 325k |
| | KNACK-2002 | (Hoste and De Pauw 2006) | 125k |
| **Catalan** | ACORA-CO-CA | (Recasens and Martí 2010) | 400k |
| **Czech** | PDT 2.0 | (Böhmová, Hajic et al. 2003) | 800k |
| **Chinese** | ACE-2005 | (Doddington, Mitchell et al. 2004) | 200k |
| | ACE-2007 | - | 250k |
| | ONTONOTES 5.0 | (Pradhan, Hovy et al. 2007) | 820k |
| **Arabic** | ACE-2005 | (Doddington, Mitchell et al. 2004) | 200k |
| | ACE-2007 | - | 250k |
| | ONTONOTES 5.0 | (Pradhan, Hovy et al. 2007) | 820k |
| **Persian** | PerCoref | (Mirzaei and Safari 2018) | 200K |
| | RCDAT | (Rahimi and HosseinNejad 2020) | 1000K |
| | Mehr | (Haji Mohammadi, Talebpour et al. 2023) | 250k |

## 3. Evaluation metrics

Evaluating the effectiveness of coreference resolution (CR) systems is crucial for guiding research and development. Various evaluation metrics have been proposed, each with its strengths and weaknesses. This section provides a comprehensive review of well-known CR metrics, examining their underlying principles, advantages, and disadvantages. Each metric assesses system performance by comparing the predicted coreference chains (response) against the gold-standard annotations (key), typically using precision and recall to compute an F1 score.

### 3.1. MUC metric

Introduced during the Sixth Message Understanding Conference (MUC-6) (Grishman and Sundheim 1996), the MUC metric is a link-based metric that compares links within coreference chains. *Recall* is defined as the proportion of correct links in the key chain that are also present in the response chain, while precision is the proportion of correct links in the response chain that are also in the key chain. Its purpose is to compare key and response chain links. In this metric, *recall* is *the number of links that must be removed from the key chain to obtain a response chain*.

In contrast to the recall metric, the *precision* metric is defined as *the number of links that must be removed from the response chain to obtain the key chain*. For a response partition based on the key defined as a partition *(r, k)* and a key partition based on the response as a partition (k, r), the precision and recall are calculated according to the following equations.

$$Precision(R, K) = \sum_{r \in R} \frac{|r| - |Partition(r,k)|}{|r| - 1} \quad (1)$$

$$Recall(K, R) = \sum_{k \in K} \frac{|k| - |Partition(k,r)|}{|k| - 1} \quad (2)$$

**Advantages:**

- Simplicity: The MUC metric is relatively easy to compute and interpret.

**Disadvantages:**

- Ignores Singletons: This metric cannot evaluate systems identifying singletons (mentions not belonging to any chain).
- Limited Discriminative Power: It may not differentiate well between systems with varying levels of accuracy (Bagga and Baldwin 1998, Luo 2005).
- Bias Towards Larger Chains: It can favor systems that produce fewer, larger chains, even if they are less accurate overall.

### 3.2. B³ metric

The $B^3$ metric (Bagga and Baldwin 1998) addresses some of the MUC metric's shortcomings by evaluating the coreference decision for each mention individually. Precision and recall are calculated for each mention, and the final scores are weighted averages across all mentions.

In this metric, recall is defined as *mapping each mention in the key to one of the response mentions, followed by calculating the number of overlapping mentions in the key and response.* The definition of precision is exactly the opposite of recall. The precision and recall for the $i_{th}$ entity are calculated according to the following expressions.

$$Precision = \frac{The\ number\ of\ correct\ elements\ in\ the\ entity\ of\ mention_i}{The\ number\ of\ mentions\ in\ response\ entity\ of\ mention_i} \quad (3)$$

$$Recall = \frac{The\ number\ of\ correct\ elements\ in\ the\ entity\ of\ mention_i}{The\ number\ of\ mentions\ in\ key\ entity\ of\ mention_i} \quad (4)$$

Final recall and precision are calculated using the following two formulas.

$$Final\ precision = \sum_{i=1}^{N} wi * precision \tag{5}$$

$$Final\ recall = \sum_{i=1}^{N} wi * recall \tag{6}$$

**Advantages:**

- Mention-Level Evaluation: Provides a more fine-grained analysis of system performance.
- Handles Singletons: Unlike MUC, $B^3$ can evaluate systems that identify singletons.

**Disadvantages:**

- Sensitivity to Repetition: Repeating mentions in the response can artificially inflate precision and recall.
- Potential Bias: If a system predicts all mentions as singletons or all mentions as belonging to a single cluster, it can achieve 100% recall or precision, respectively.

### 3.3. CEAF metric

The CEAF metric (Luo 2005) is an entity-based metric that aims to identify similarities between key and response entities. The mapping function is defined with a scoring function $\varphi\ (g)$:

$$\varphi(g) = \sum_{k \in k_m} \varphi(k, g(k)) \tag{7}$$

Given document *D* and the key and response entities, the optimal mapping that maximizes total similarity can be determined as follows:

$$g^* = \underset{g \in G_m}{\arg\max}\ \varphi(g) \tag{8}$$

Then:

$$\varphi(g^*) = \sum_{k_i\ k^*} \varphi(k_i, g^*(k_i)) \tag{9}$$

After calculating $\varphi\ (k, k)$ and $\varphi\ (r, r)$, precision and recall are calculated as follows:

$$Precision = \frac{\varphi(g^*)}{\sum_{r_i\ \in r^*} \varphi(r_i, r_i)} \tag{10}$$

$$Recall = \frac{\varphi(g^*)}{\sum_{k_i\ \in k^*} \varphi(k_i, k_i)} \tag{11}$$

The similarity metric can be calculated using the following four equations. The third is used in $CEAF_m$, and the fourth is used to calculate $CEAF_e$.

$$\varphi_1(K,R) = \begin{cases} 1 \text{ if } R = K \\ 0 \text{ otherwise} \end{cases} \tag{12}$$

$$\varphi_2(K,R) = \begin{cases} 1 \text{ if } R \cap K \neq \emptyset \\ 0 \text{ otherwise} \end{cases} \tag{13}$$

$$\varphi_3(K,R) = |R \cap K| \tag{14}$$

$$\varphi_4(K,R) = 2 * \frac{|R \cap K|}{|R|+|K|} \tag{15}$$

**Advantages:**

- Focus on Entities: Aligns with the core goal of CR, which is to identify entities rather than just links.
- Flexible Similarity Functions: Different similarity functions can capture various aspects of entity matching.

**Disadvantages:**

- Complexity: The calculation can be more complex than link-based metrics.
- Potential Bias: The choice of similarity function can influence the evaluation results.

### 3.4. CoNLL metric

In 2012, during the CoNLL shared task, a metric that averaged the previous three metrics was introduced and became the standard for comparing coreference systems. The corpus utilized in this competition was OntoNotes (Pradhan, Moschitti et al. 2012). This metric is computed using the formula shown below.

$$CoNLL = \frac{(MUC_{F1} + B^3_{F1} + CEAFe_{F1})}{3} \tag{16}$$

**Advantages:**

- Comprehensive: Combines the strengths of multiple metrics.
- Widely Used: It has become a standard benchmark for comparing CR systems.

**Disadvantages:**

- May Obscure Individual Metric Weaknesses: Averaging can mask the shortcomings of any single metric.

### 3.5. ACE-Value metric (Un-normalized Entity-Based)

The ACE-Value metric, similar to CEAF, seeks an optimal mapping between key and response entities. However, it needs to normalize precision and recall, making it less common in recent research.

**Advantages:**

- **Entity-Centric Focus:** Like the CEAF metric, ACE-Value prioritizes the correct identification and grouping of entities, aligning well with the core objective of coreference resolution. It seeks the optimal mapping between key and response entities, emphasizing the accurate clustering of mentions that refer to the same real-world entity.
- **Flexibility:** ACE-Value can be adapted to various evaluation scenarios by modifying the underlying similarity function. This allows researchers to tailor the metric to specific domains or applications where certain entity matches might be more important than others.

**Disadvantages:**

- **Un-normalized Scores:** ACE-Value does not normalize precision and recall scores like CEAF. This means that the scores can be influenced by the absolute number of mentions in the key and response, making it difficult to compare results across different datasets or systems with varying sizes.
- **Less Commonly Used:** Due to the lack of normalization and the emergence of more sophisticated metrics like CEAF and CoNLL, ACE-Value is not as widely used in recent coreference resolution research.
- **Potentially Less Informative:** he un-normalized nature of ACE-Value may provide less detailed information about system performance compared to metrics like $B^3$, which offers mention-level precision and recall.

### 3.6. BLANC metric

This link-based metric was introduced by (Recasens and Hovy 2011). Additionally, it is an adaptation of the Rand index (Rand 1971) for coreference.

Suppose $C_K$, $C_R$, $N_k$, and $N_R$ denote coreference chains in the key, coreference chains in response, non-coreference chains in the key, and non-coreference chains in response. In that case, the precision and recall for this metric can be calculated using the following expressions.

$$R_C = \frac{|C_K \cap C_R|}{|C_K|} \tag{17}$$

$$R_N = \frac{|N_K \cap N_R|}{|N_K|} \tag{18}$$

$$P_C = \frac{|C_K \cap C_R|}{|C_R|} \tag{19}$$

$$P_N = \frac{|N_K \cap N_R|}{|N|} \quad (20)$$

$$Precision = \frac{P_C + P_N}{2} \quad (21)$$

$$Recall = \frac{R_C + R_N}{2} \quad (22)$$

In a baseline system where all mentions are connected and form a single entity, this metric performs poorly against the $B^3$ and CEAF metrics. The identification effects mentioned above are significant in this metric (Moosavi and Strube 2016) due to the utilization of non-coreference relations. As a result, it is less prevalent in relevant works.

**Advantages:**

- Considers Non-Coreference: Unlike other link-based metrics, it accounts for non-coreferent mentions, potentially offering a more nuanced evaluation.

**Disadvantages:**

- Sensitivity to Mention Identification: This may be affected by the number of mentions identified, regardless of their accuracy (mention identification effect).
- Less Common: Not as widely used as other metrics due to its potential biases.

### 3.7. LEA metric (Link-Based, Entity-Aware)

This metric was proposed by (Moosavi and Strube 2016) to counteract the previously mentioned identification effect. The mention identification effect for a metric is that if the system finds more mentions, it will increase system performance, regardless of whether they are true or false. The authors artificially added links to their system, which are in the key but not the response. Consequently, an increase in $B^3$, CEAF, and BLANK metrics is observed. The only metric capable of overcoming this issue is the MUC metric, which has other issues described previously.

This metric is based on two terms: importance and resolution score. The first one depends on the size of the corpus, and the second one is calculated based on the links' similarity. In this metric, chains receiving more mentions will receive higher scores. According to (Luo 2005), discriminative power and interpretability are two reasonable prerequisites for coreference systems that CoNLL metrics fail to meet.

In this metric, the authors consider an entity's size. To circumvent the problem of the mention identification effect, they evaluate the coreference relationships between entities rather than the resolved mentions. Consequently, duplicate mentions in response entities will not affect this metric. Since it considers all links, its separation power will be greater than MUC's.

Resolution-score for the $k_{ith}$ key entity is calculated as a fraction of correctly co-referenced links.

$$Res\_score(K_i) = \sum_{r_j \in R} \frac{link(K_i \cap r_j)}{link(K_i)} \quad (23)$$

Precision and recall are calculated as follows:

$$Importance(e_i) = |e_i| \quad (24)$$

$$Precision = \frac{\sum_{r_i \in R} Importance(r_i) * \sum_{k_j \in K} \frac{link(r_i \cap k_j)}{link(r_i)}}{\sum_{r_z \in R} Importance(r_z)} \quad (25)$$

$$Recall = \frac{\sum_{k_i \in K} Importance(K_i) * \sum_{r_j \in R} \frac{link(k_i \cap r_j)}{link(k_i)}}{\sum_{r_z \in K} Importance(k_z)} \quad (26)$$

**Advantages:**

- Mitigates Mention Identification Effect: Less sensitive to the number of mentions identified.
- Higher Discriminative Power Considers all links, offering better system differentiation.

**Disadvantages:**

- Complexity: The calculation is relatively complex compared to other metrics.

### 3.8. NEC metric (Mention Type-Aware)

(Agarwal, Subramanian et al. 2019) introduced the Named Entity Coreference (NEC) metric. This metric considers the practical applications of coreference resolution in the subsequent task. In NEC, the type of mentions is considered in the final f1-score. The following is an example from the original article:

**Gold chain = [John Doe, he1, he2, he3] [Richard Roe, he4, he5]**

**System output 1 = [John Doe, he1, he2] [Richard Roe, he4, he5]**

**System output 1 = [he1, he2] [he4, he5]**

Because current evaluation metrics do not account for the type of mentions, both outputs receive the same score. The first output receives higher NEC scores because the first system's output refers to a nominal mention. The NEC metric calculation is as follows.

For every $k_i \in K$, assume $N_i$ is the set of response mentions with a full name of $k_i$.

$$Precision = p(k_i, r_j) = \frac{|r_j \cap k_i|}{|r_j|} \quad (27)$$

$$Recall = r(k_i, r_j) = \frac{|r_j \cap k_i|}{|k_i|} \quad (28)$$

The f1 value is then calculated for the response and key entities as follows:

$$f(k_i, r_j) = r(k_i, r_j) = \frac{2p(k_i,r_j)r(k_i,r_j)}{r(k_i,r_j)+ p(k_i,r_j)} = 2 \frac{|r_j \cap k_i|}{|k_i|+ |r_j|} \tag{29}$$

Then, the following value is assigned to f1 for the $k_i$ entity:

$$F1_i = \max f(k_i, r_j)_{r_j \in R:\, r_j \cap N_i \neq \emptyset} \tag{30}$$

Finally, the f1 value for the whole system is as follows:

$$\frac{1}{|K|} \sum_{k_i \in K} F1_i \tag{31}$$

**Advantages:**

- Application-Oriented: Aligns with real-world CR use cases where named entity resolution is often critical.

**Disadvantages:**

- Limited Scope: This may only be suitable for some CR tasks, particularly those not focused on named entities.

### 3.9. A simple metric comparison example

In this section, we will review an example adopted from Pradhan et al. (Pradhan, Luo et al. 2014). Suppose there are two chains in the key *"K"*: [a, b, c] and [d, e, f, g]. In response *"R"*, there are three chains [a, b], [c, d], and [f, g, h, i].

$$K = [a, b, c], [d, e, f, g]$$

$$R = [a, b], [c, d], [f, g, h, i]$$

Precision and recall for the metric above are examined below.

#### 3.9.1. MUC metric

Considering the chains above, calculating precision and recall is straightforward.

$$\text{Precision}(R, K) = \sum_{r \in R} \frac{|r| - |\text{Partition}(r, k)|}{|r| - 1}$$

$$= \frac{(3-2) + (4-3)}{(3-1) + (4-1)} = 0.4$$

$$\text{Recall}(K, R) = \sum_{k \in K} \frac{|k| - |\text{Partition}(k, r)|}{|k| - 1}$$

$$= \frac{(2-1) + (2-2) + (4-3)}{(2-1) + (2-1) + (4-1)} = 0.4$$

### 3.9.2. $B^3$ metric

Based on the chains above, precision and recall are as follows:

$$\text{Recall} = \frac{1}{7} * (\frac{2^2}{3} + \frac{1^2}{3} + \frac{1^2}{4} + \frac{2^2}{4}) \approx 0.42$$

$$\text{Precision} = \frac{1}{8} * (\frac{2^2}{2} + \frac{1^2}{2} + \frac{1^2}{2} + \frac{2^2}{4}) = 0.5$$

### 3.9.3. CEAF metric

The $\text{CEAF}_m$ is calculated according to the similarity metric in Eq. (14). The value of $\text{CEAF}_m$ $f_1$ is 0.53.

$$\text{Recall} = \frac{|R_1 \cap K_1| + |R_3 \cap K_2|}{|K_1| + |K_2|} = \frac{2+2}{3+4} \approx 0.57$$

$$\text{Precision} = \frac{|R_1 \cap K_1| + |R_3 \cap K_2|}{|R_1| + |R_2| + |R_3|} = \frac{2+2}{2+2+4} = 0.5$$

The $\text{CEAF}_e$ metric is then computed. Using Eq. (15), the similarity metric for this metric is as follows.

$$\varphi_4(K, R) = 2 * \frac{|R \cap K|}{|R| + |K|}$$

The value of $\text{CEAF}_e$ $f_1$ is 0.52.

$$\text{Recall} = \frac{\varphi_4(K_1, R_1) + \varphi_4(K_2, R_3)}{N_k} = \frac{\frac{2*2}{3+2} + \frac{2*2}{4+4}}{2} = 0.65$$

$$\text{Precision} = \frac{\varphi_4(K_1, R_1) + \varphi_4(K_2, R_3)}{N_r} = \frac{\frac{2*2}{3+2} + \frac{2*2}{4+4}}{3} \approx 0.43$$

*3.9.4. BLANK metric*

For this metric, first, the values of $C_k$, $N_k$, $C_r$, and $N_r$ are computed according to Eqs. (17-20).

$$C_k = \{ab, ac, bc, de, df, dg, ef, eg, fg\}$$

$$N_k = \{ad, ae, af, ag, bd, be, bf, bg, cd, ce, cf, cg\}$$

$$C_r = \{ab, cd, fg, fh, fi, gh, gi, hi\}$$

$$N_r = \{ac, a, af, ag, ah, ai, bc, bd, bf, bg, bh, bi, cf, cg, ch, ci, df, dg, dh, di\}$$

Precision and recall are as follows for both coreference and non-coreference links:

$$R_C = \frac{|C_k \cap C_r|}{|C_k|} = \frac{2}{9} \approx 0.22$$

$$P_C = \frac{|C_k \cap C_r|}{|C_r|} = \frac{2}{8} = 0.25$$

$$R_n = \frac{|N_k \cap N_r|}{|N_k|} = \frac{8}{12} \approx 0.67$$

$$P_n = \frac{|N_k \cap N_r|}{|N_r|} = \frac{8}{20} = 0.4$$

According to Eqs. (21) and (22), $F_C$ is approximately 0.23, and $F_n$ is 0.5. The BLANK will be an average of $F_c$ and $F_n$.

*3.9.5. LEA metric*

In this example, the set $S$ is a key set and consists of two chains $s_1$ = [a, b, c] and $s_2$ = [d, e, f, g], and the set $R$ is a response set and comprises three chains $r_1$ = [a, b], $r_2$ = [c, d] and $r_3$ = [f, g, h, i]. The importance value of an entity is equal to its size. Therefore, it is 3 for chain $s_1$ and 4 for chain $s_2$. The set of coreference links in $s_1$ and $s_2$ are [ab,ac,bc], and [de,df,dg,ef,eg,fg], respectively.

The *"ab"* link is the only common connection between $s_1$ and $r_1$; no common link exists between $s_1$ and the other two responses. Cluster *s2* has a common link with *r3* and lacks a common link with the other two response sets. The resolution-score value for chains *s1* and *s2* is calculated according to Eq. (23).

$$\text{Res\_score}(K_1) = \frac{1 + 0 + 0}{3}$$

$$\text{Res\_score}(K_2) = \frac{0 + 0 + 1}{6}$$

Similar calculations are made for the importance and resolution-score values of the response clusters. Next, the precision and recall values are calculated through Eqs. (25) and (26).

$$\text{Precision} = \frac{2 * \frac{1+0}{1} + 2 * \frac{0+0}{1} + 4 * \frac{0+1}{6}}{2+2+4} \approx 0.33$$

$$\text{Recall} = \frac{3 * \frac{1}{3} + 4 * \frac{1}{6}}{3+4} \approx 0.24$$

### *3.10. Comparative Analysis and Future Directions*

The choice of evaluation metric can significantly impact the assessment of CR system performance. While CoNLL has become a standard, its reliance on averaging multiple metrics may obscure individual weaknesses. LEA's focus on entity-level evaluation and mitigation of the mention identification effect makes it a promising alternative for specific tasks. The NEC metric, on the other hand, caters to specific applications where named entity resolution is paramount.

Future research should aim to develop evaluation metrics that are even more aligned with real-world CR applications. This may involve incorporating semantic similarity measures, discourse coherence, or task-specific performance. Additionally, more standardized benchmarks and datasets are needed to facilitate fair and comprehensive comparisons across different CR systems, especially for underrepresented languages.

By developing more robust and nuanced evaluation metrics, coreference resolution can continue to advance towards systems that more accurately and reliably understand the complex relationships between entities and events in natural language text.

## 4. Coreference Algorithms: From Handcrafted Rules to Neural Networks

As detailed in numerous articles and books, the computational landscape of coreference resolution (CR) has undergone a remarkable evolution. Early CR systems primarily relied on hand-crafted linguistic rules and domain-specific knowledge. However, a paradigm shift occurred in the late 1990s, with machine learning methods gradually replacing rule-based approaches. More recently, the field has witnessed a surge in hybrid and deep learning systems, pushing the boundaries of CR performance. This section embarks on a journey through the diverse methodologies employed in CR, starting with the foundational rule-based systems and culminating in state-of-the-art neural network models. We will also focus on anaphora and coreference resolution systems developed for the Persian language.

### 4.1 Rule-based systems

The nascent era of coreference resolution (CR) was characterized by systems that relied on hand-crafted linguistic rules and heuristics. These systems, often the product of meticulous linguistic analysis and domain expertise, aimed to emulate human understanding of coreference through explicit knowledge encoding.

Early systems, like Hobbs' pronoun resolution algorithm (Hobbs 1978), utilized syntactic parse trees as a guiding principle, underscoring the importance of structural knowledge in identifying antecedents. While these pioneering efforts laid the groundwork for CR research, their reliance on manual evaluation and deep linguistic analysis could have improved their scalability and generalizability. A pivotal shift occurred with the advent of salience-based models, which emphasized the role of discourse prominence in antecedent selection (Sidner 1979, Carter 1987). This approach, combined with syntactic and semantic features integration, as seen in (Lappin and Leass 1994), improved performance on larger corpora.

A key distinction emerged between knowledge-rich and knowledge-lean rule-based systems. Knowledge-rich systems, like (Lappin and Leass 1994), leveraged in-depth linguistic knowledge and complex parsers, often achieving high accuracy at the expense of generalizability and computational efficiency. Conversely, knowledge-lean systems, exemplified by CogNIAC (Baldwin 1997) and MARS (Mitkov 1998), favored practicality and scalability. These systems relied on simpler heuristics and surface-level features, facilitating their adaptation to diverse domains and languages.

The Stanford CoreNLP system (Lee, Peirsman et al. 2011), a testament to the effectiveness of rule-based approaches, emerged as the winner of the CoNLL-2011 shared task. Its architecture, composed of a cascade of 12 coreference sieves, each with hand-crafted rules targeting specific linguistic phenomena, showcased the power of a modular and iterative approach to rule-based CR.

Rule-based systems have left an enduring legacy in CR research, providing valuable insights into the linguistic factors that govern coreference phenomena. However, their reliance on manual rule crafting presents several challenges that ultimately limit their effectiveness and applicability in real-world scenarios:

- **Brittleness and Lack of Robustness:** Hand-crafted rules are often brittle and need help handling natural language's inherent variability and ambiguity. Even minor deviations from the anticipated linguistic patterns can lead to errors in coreference resolution.
- **Limited Generalizability:** Rules designed for one domain or language may not transfer to others. The diversity of linguistic structures and discourse conventions across languages and genres necessitates the development of specialized rule sets, making it challenging to create a universally applicable rule-based CR system.
- **High Development and Maintenance Costs:** Crafting, refining, and maintaining a comprehensive set of rules for CR is labor-intensive and time-consuming. It requires expert linguistic knowledge and continuous updates to keep up with evolving language use.

- **Difficulty in Handling Ambiguity:** Rule-based systems often struggle to resolve ambiguous references where multiple antecedents are plausible. This is particularly challenging for pronouns, which can have diverse referents depending on context.
- **Subjectivity and Bias:** Hand-crafted rules can inadvertently introduce biases and subjectivity into the CR process. The rule developer's linguistic intuitions and assumptions can shape the system's behavior, potentially leading to biased or inconsistent results.
- **Inability to Learn from Data:** Unlike machine learning-based approaches, rule-based systems cannot learn from data and improve their performance over time. This makes them less adaptable to new information and less effective in handling complex or noisy real-world data.

In conclusion, rule-based coreference resolution systems played a foundational role in the development of CR research, offering valuable insights into linguistic factors and coreference phenomena. However, their limitations, including brittleness, limited generalizability, high development costs, and difficulty in handling ambiguity, paved the way for the emergence of data-driven approaches. Table 2 provides an overview of various rule-based systems, highlighting evaluation methodologies across different datasets and metrics.

**Table 2.** Comparison of rule-based systems

| Algorithm | Dataset | Evaluation Metric | Metric Score |
|---|---|---|---|
| (Charniak 1972) | Manually on a small scale | Not systematically evaluated | - |
| (Hobbs 1978) | Narrative texts | Hobb's metric | 88.3 |
| (Brennan, Friedman et al. 1987) | Narrative texts | Hobb's metric | 90 |
| (Lappin and Leass 1994) | Five Computer Science manuals | Hobb's metric | 89 |
| (Kennedy and Boguraev 1996) | News text | Resolution accuracy | 75 |
| (Baldwin 1997) | MUC-6 | Precision and Recall | P:73, R:75 |
| (Kameyama 1997) | MUC-6 | MUC $F_1$ | 65 |
| (Mitkov 1998) | Various (English, Arabic, Polish, Bulgarian) | MUC, B3, CEAF | Varies by dataset |
| (Gaizauskas, Wakao et al. 1995) | MUC-6 | MUC $F_1$ | 59 |
| (Tetreault 2001) | News articles | Hobb's metric | 80.4 |
| (Lee, Peirsman et al. 2011) | MUC-6 | MUC, $B^3$ | MUC: 77.7, $B^3$:73.2 |
| (Lee, Chang et al. 2013) | CoNLL 2012 | CoNLL | 60.13 |

### 4.2. Early machine learning models (non-neural systems)

The emergence and expansion of large-scale coreference corpora and advances in machine learning (ML) architectures spurred a paradigm shift in coreference resolution (CR) research. Rule-based systems, while insightful, struggled to resolve pronouns accurately due to the inherent difficulty of capturing complex contextual information through hand-crafted rules. Machine learning, in contrast, offered the promise of automatically extracting intricate feature interactions from data, given sufficient training examples. This ushered in an era of data-driven CR models, which we explore in detail in the following section.

*4.2.1. Mention-pair model*

The mention-pair model, pioneered by (Aone and Bennett 1995) and solidified by (Soon, Ng et al. 2001), represents a foundational approach in machine learning-based coreference resolution (CR). This model frames CR as a binary classification task, where pairs of noun phrases (mentions) are evaluated for coreference using a feature-based representation.

The mention-pair model comprises three distinct components:

- **Training Sample Generation:** This stage involves constructing positive (coreferent) and negative (non-coreferent) mention pairs from the corpus. Addressing class imbalance is often crucial, with standard techniques involving sampling strategies or modifying the selection of negative instances (Harabagiu, Bunescu et al. 2001, Ng and Cardie 2002, Yangy, Su et al. 2004). Recent work has explored more sophisticated sampling methods, such as those based on active learning (Li, Stanovsky et al. 2020) or hard negative mining, to improve training data's balance and quality.
- **Classifier Training:** Various classifiers have been employed to learn the underlying patterns of coreference from the training data. Popular choices include decision trees (Soon, Ng et al. 2001, Ng and Cardie 2002, Lee, Surdeanu et al. 2017), maximum entropy models(Luo, Ittycheriah et al. 2004, Ng 2005, Nicolae and Nicolae 2006, Ponzetto and Strube 2006), RIPPER rule-based learners (Ng and Cardie 2002, Hoste 2005, Ng 2005), statistical methods (Ge, Hale et al. 1998), and support vector machines (SVMs) (Uryupina 2006, Rahman and Ng 2009). Recent research has focused on leveraging deep learning models, such as the end-to-end neural coreference resolution model (Lee, He et al. 2017), to learn rich feature representations for mention-pair classification and ranking automatically.
- **Coreference Clustering:** After pairwise decisions are made, various clustering strategies are employed to group coreferent mentions into chains. Common approaches include link-first (Soon, Ng et al. 2001), best-first (Ng and Cardie 2002, Bengtson and Roth 2008), correlation clustering (McCallum and Wellner 2005), Bell tree (Luo, Ittycheriah et al. 2004), and graph clustering (Nicolae and Nicolae 2006). Recent work has explored more sophisticated clustering methods, such as agglomerative clustering with learned distance metrics (Kantor and Globerson 2019) and end-to-end neural models that jointly learn mention representations and perform clustering (Lee, He et al. 2017) to improve coreference chains' accuracy and coherence.

Despite its simplicity and reliance on surface features, the mention-pair model has demonstrated competitive performance compared to early rule-based systems. Its modular design allows for independent improvements in each component and has served as a basis for numerous successful systems in CoNLL evaluations. Early attempts to incorporate semantic knowledge into these models explored features derived from resources like WordNet, with varying degrees of success (Huang, Zeng et al. 2009).

However, this model has several limitations. Notably, its pairwise nature leads to local optimality issues, where individual decisions may lead to something other than globally optimal coreference chains. For example, if (Bill Clinton, Clinton) and (Clinton, She) are both classified as coreferent, an erroneous chain (Bill Clinton, Clinton, She) could be formed due to transitivity.

Additionally, the mention-pair model only assesses the suitability of a single antecedent candidate for a given mention without considering the relative merits of other candidates. This limitation and the need for extensive training sets to combat class imbalance motivated the development of alternative models, such as entity-based and ranking models.

### *4.2.2. Entity-based model (clustering model)*: *A Global Perspective on Coreference*

Unlike mention-pair models, entity-based models leverage prior knowledge, particularly partial entity information, to resolve coreference. They frame CR as an incremental clustering problem rather than a series of binary classification decisions. This allows for a holistic assessment of coreference relationships by directly comparing mentions to evolving clusters of potential referents.

One of the earliest entity-based models was introduced, employing a cluster merging operation that checked compatibility between mentions across clusters (Cardie and Wagstaf 1999). However, their work did not directly compare this approach to the mention-pair model. Graph partitioning (Nicolae and Nicolae 2006, Culotta, Wick et al. 2007, Cai and Strube 2010) emerged as a prominent technique within the entity-based paradigm. In this model, mentions are represented as nodes in a graph, and edges between nodes signify potential coreference relationships. The edges are weighted based on binary or unary features extracted from the mentions. A graph-cutting algorithm then partitions the graph based on these weights, forming coreference clusters (Figure 2).

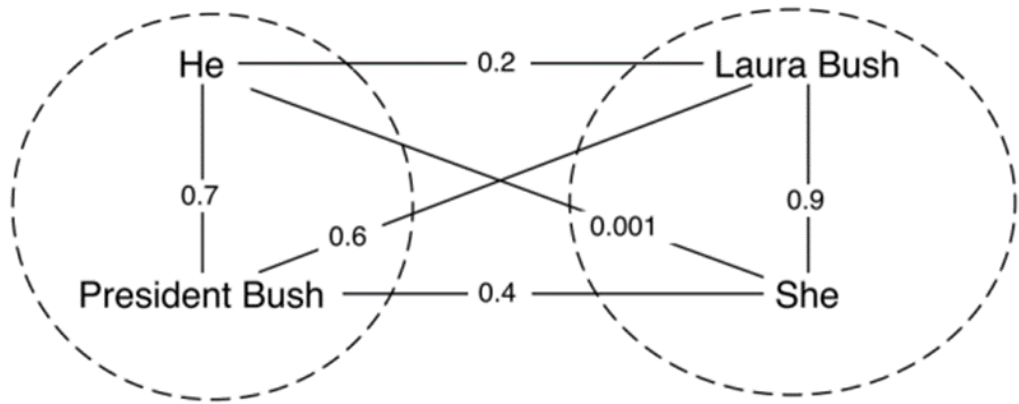

Figure 2. Graph-based modeling of coreference resolution. The circles within the image represent a concentration of the golden corpus. The image is from Culotta et al. (Culotta, Wick et al. 2007)

(Nicolae and Nicolae 2006) utilized the "best-cut" method to prune edges in the graph, demonstrating superior performance to entity-mention models while not employing cluster-based features (Luo, Ittycheriah et al. 2004). (Cai and Strube 2010) further advanced this approach by introducing the concept of hyper-edges, leading to one of the top-performing systems in the CoNLL-2011 shared task. (Culotta, Wick et al. 2007) employed first-order logic to define cluster-level properties, significantly outperforming mention-pair models on the ACE corpus. Their approach offered the advantage of resolving coreference in a single step and considering multiple links simultaneously, though it did not explicitly isolate the impact of cluster-based features.

To address inconsistencies in mention-pair model clustering, researchers integrated integer linear programming (ILP) as an additional component (Finkel and Manning 2008, Klenner and Ailloud 2008, Denis and Baldridge 2009, Klenner and Ailloud 2009) ILP optimizes the clustering process by receiving pairwise decision weights as input and ensuring uniqueness and transitivity in coreference chains. However, this method incurs significant computational costs and relies on the underlying mention-pair mechanism.

A clustering method prevents inconsistencies between coreference chains when pairing mention-pair model samples. In this system, an additional component, such as integer linear programming (ILP), ensures uniqueness and Transitive closure when pairing the mention pairs (Finkel and Manning 2008, Klenner and Ailloud 2008, Denis and Baldridge 2009, Klenner and Ailloud 2009). As input, the ILP layer receives the weights of the pair decisions and optimizes the clustering operation. However, many engineering mechanisms and computational efforts are required for this method. One of its issues is its continued reliance on the mechanism that generates mention-pairs. ILP systems are NP-hard problems, which is another drawback of this method.

**Advantages:**

- **Global Coherence:** Entity-based models prioritize the overall coherence of coreference clusters, considering the compatibility of multiple mentions within a cluster rather than just pairwise relationships. This can lead to more accurate and consistent resolutions, especially for complex documents with multiple entities.
- **Reduced Cascading Errors:** By evaluating mentions against entire clusters, entity-based models can mitigate the cascading error problem often seen in mention-pair models, where incorrect pairwise decisions can propagate through the resolution process.
- **Incorporating Entity-Level Features:** These models can easily incorporate features that capture properties of entire entities (e.g., gender, number, semantic coherence), providing richer information for coreference decisions.

**Disadvantages:**

- **Computational Complexity:** Entity-based models can be computationally expensive, especially when dealing with large documents or complex clustering algorithms. Considering all pairwise relationships within a cluster can lead to increased processing time.
- **Sensitivity to Initialization:** The performance of some clustering algorithms can be sensitive to the initial cluster assignments. Poor initialization can lead to suboptimal solutions and affect the overall accuracy of coreference resolution.
- **Difficulty in Handling Ambiguity:** While entity-based models consider global context, they may still need help with ambiguous references where multiple antecedents are plausible. This is particularly challenging when limited information is available within the existing clusters.

*4.2.3. Entity-mention model*

Entity-mention models occupy a unique position in the landscape of early machine learning approaches to coreference resolution (CR). They bridge the gap between mention-pair and entity-based models by considering pairwise decisions and partial entity information. This enables a more comprehensive evaluation of coreference relationships than purely pairwise approaches.

In the entity-mention model, pairwise decisions between mentions are not discarded but are stored and utilized in subsequent decisions. Instead of solely examining pairs of mentions, the model evaluates the compatibility of a given mention with previously formed partial coreference chains (entities). This allows for incorporating local (pairwise) and global (entity-level) information into the decision-making process. The model operates by incrementally forming coreference chains as it processes the text. The model considers its potential coreference relationships with existing chains for each new mention. It computes a feature vector representing the mention and the chain, incorporating both mention-level and cluster-based features. A classifier then determines whether the mention should be added to the chain, creating a new chain, or remaining as a singleton.

(Luo, Ittycheriah et al. 2004) introduced a notable entity-mention model that combines global optimization of coreference chains. They represented partial clusters of mentions as a tree structure called a Bell tree (Figure 3). The model determines whether each mention should join an existing chain or start a new one by traversing the tree and evaluating feature scores between the mention and each potential antecedent chain.

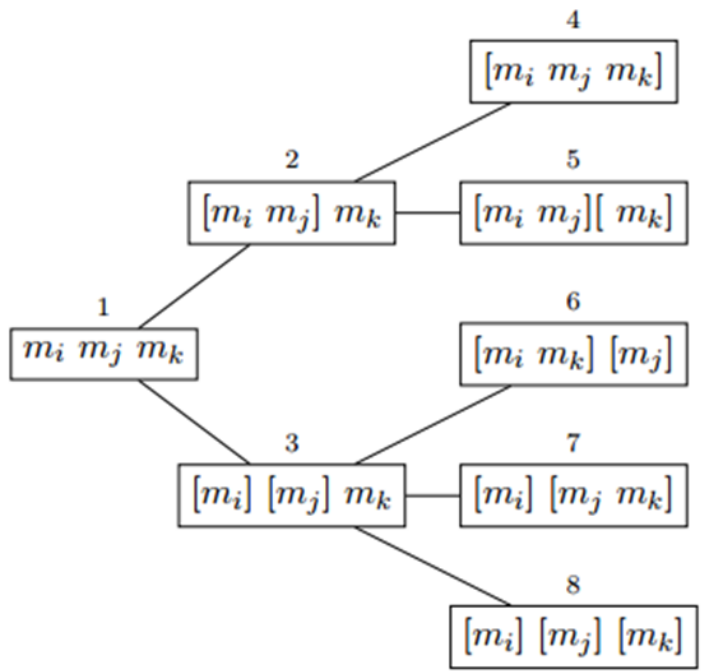

Figure 3. Example of a Bell tree adopted from Luo et al. for three mentions (Luo, Ittycheriah et al. 2004)

While their model did not outperform mention-pair models on the ACE corpus, it demonstrated the potential of utilizing fewer features while still achieving reasonable results. However, their approach also introduced inconsistencies, such as potentially grouping pronouns with different genders into the same chain.

(Yangy, Su et al. 2004) adapted the entity-mention model for medical texts, iteratively forming chains and comparing each mention to preceding partial chains. Their evaluation showed the effectiveness of the entity-mention model in improving CR performance. Later, (Yang, Su et al. 2008) further enhanced this approach by using inferential logic programming to learn coreference rules from data, demonstrating the flexibility and potential of this model.

(Daumé III and Marcu 2005) proposed a unique entity-mention model that jointly performed mention detection and coreference resolution. Their system utilized a rich feature set but needed help accurately identifying pronoun references, highlighting the challenges of effectively integrating these two tasks. Entity-mention models offer several advantages over mention-pair models, such as leveraging global information from partial entities and reducing cascading errors. However, they can be computationally expensive due to the need to evaluate mentions against multiple chains, and they may still need more clarity when limited information is available within the existing clusters. Despite these challenges, entity-mention models have significantly contributed to the advancement of CR research, paving the way for more sophisticated models that integrate local and global information for more accurate and robust coreference resolution.

### *4.2.4. Ranking systems: A Flexible Approach for Prioritizing Antecedents*

Ranking models offer a flexible compromise between the local focus of mention-pair models and the global perspective of entity-based models. They enable the simultaneous consideration of multiple antecedent candidates for a given mention, prioritizing the most likely choice based on learned features.

The concept of ranking for coreference resolution was first introduced by (Connolly, Burger et al. 1997), who proposed a system for ranking two mention candidates. This was followed by the twin-candidate model(Yang, Zhou et al. 2003, Denis and Baldridge 2008), which expanded the ranking to multiple antecedent candidates, eliminating the need for explicit clustering. (Durrett and Klein 2013) proposed a mention ranking model that relied solely on shallow features, demonstrating competitive performance and efficiency compared to the Stanford system (Raghunathan, Lee et al. 2010, Lee, Peirsman et al. 2011, Lee, Chang et al. 2013) and the structured perceptron model (Fernandes, Dos Santos et al. 2012).

**Advantages**

- **Global Context:** Ranking models can consider multiple antecedent candidates simultaneously, considering a more comprehensive range of contextual information than mention-pair models.
- **Flexibility:** The ranking approach allows for incorporating diverse features, including those derived from syntactic, semantic, and discourse analysis.
- **Efficiency:** Ranking models can be computationally efficient, as they do not require explicit clustering steps.

**Disadvantages**

- **Local Optimality:** While ranking models consider multiple candidates, they still operate at the mention-pair level and may not guarantee globally optimal solutions.
- **Lack of Entity-Level Information:** These models traditionally have yet to leverage entity-level information, which could be beneficial for resolving more complex coreference relationships.

Recent ranking-based deep learning models (Wiseman, Rush et al. 2015, Clark and Manning 2016, Wiseman, Rush et al. 2016) have successfully distinguished referential from non-referential mentions, further enhancing the accuracy and applicability of ranking models in CR. This model continues to have the issue of mention-pair local decisions. Since information is not aggregated at the entity level in this model, it will not affect subsequent decisions. This issue prompted the development of the cluster-ranking model. The cluster ranking model combines entity-based and ranking models' beneficial characteristics. Clark and Manning's (Clark and Manning 2016) deep neural network model is an example of a cluster ranking model. Table 3 compares early systems for machine learning.

Ranking models have played a vital role in advancing the field of CR, offering a flexible and efficient approach for prioritizing antecedent candidates. Recent advancements in cluster-ranking and deep learning have further enhanced their capabilities, addressing limitations and pushing the boundaries of CR performance. As research continues to evolve, we can expect even more sophisticated ranking models that leverage richer contextual information and better capture the nuances of coreference in natural language.

Table 3. Comparison of early coreference machine learning systems

| Algorithm | Learning Algorithm | Dataset | Evaluation Metric | Metric Score |
|---|---|---|---|---|
| Mention-Pair model | | | | |
| (McCarthy and Lehnert 1995) | C4.5 | MUC-6 | MUC | 47.2 |
| (Soon, Ng et al. 2001) | C4.5 | MUC-6 | MUC | 62.6 |
| (Ng and Cardie 2002) | C4.5 | MUC-6 | MUC | 69.1 |
| (Ng and Cardie 2002) | RIPPER | MUC-6 | MUC | 70.4 |
| (Yang, Zhou et al. 2003) | C5.0 | MUC-6 | MUC | 78.3 |
| (McCallum and Wellner 2005) | CRF hidden markov | MUC-6 | MUC | 73.42 |
| (Bengtson and Roth 2008) | Average perceptron | MUC | MUC | 75.8 |
| | | | $B^3$ | 80.8 |
| (Denis and Baldridge 2007) | Max Entropy | ACE-NPAPER | MUC | 72.5 |
| Entity-Mention model | | | | |
| (Luo, Ittycheriah et al. 2004) | Max Entropy | ACE | ACE-val | 89.9 |
| (Yangy, Su et al. 2004) | C5.0 | GENIA | F-measure | 81.7 |
| (Yang, Su et al. 2008) | ILP | ACE-BNews | MUC | 63.5 |
| Entity-based model (clustering model) | | | | |
| (Cardie and Wagstaf 1999) | C4.5 | MUC-6 | MUC | 54 |
| (Nicolae and Nicolae 2006) | Maximum Entropy Model | MUC-6 | MUC | 89.63 |
| (Culotta, Wick et al. 2007) | first-order logic | ACE | B3 | 79.3 |
| (Finkel and Manning 2008) | Logistic Classifier | MUC-6 | MUC | 68.3 |
| (Denis and Baldridge 2009) | Integer Linear Programming | ACE | CoNLL | 69.7 |
| (Cai and Strube 2010) | End to end clustering | MUC-6 | MUC | 64.5 |

| (Fernandes, Dos Santos et al. 2012) | Structured Perceptron | CoNLL-2012 | CoNLL | 60.65 |
|---|---|---|---|---|
| **Ranking systems** | | | | |
| (Connolly, Burger et al. 1997) | - | MUC-6 | MUC | 52.2 |
| (Yang, Zhou et al. 2003) | C4.5 | MUC-6 | MUC | 71.3 |
| (Denis and Baldridge 2008) | Max Entropy | ACE | CoNLL | 70.4 |
| (Durrett and Klein 2013) | - | CoNLL-2012 | CoNLL | 61.7 |

### 4.3. Deep learning systems: A Transformative Era in Coreference Resolution

The advent of deep learning, fueled by advances in hardware and novel architectures, revolutionized coreference resolution (CR). Early neural CR systems, while still relying on hand-crafted features for mention detection, introduced non-linear models to capture the complex relationships between mentions. This marked a crucial departure from linear classifiers and opened up new possibilities for representing linguistic knowledge in a more nuanced way.

The groundbreaking work of (Lee, He et al. 2017) marked the introduction of end-to-end neural CR systems, where mention detection and coreference resolution were integrated into a single model. This approach eliminated the need for separate pipeline modules, streamlining the process and often improving efficiency.

Deep learning CR models can be categorized into four paradigms, as illustrated in Figure 4: mention-pair, entity-mention, cluster-ranking, and end-to-end models. Each category represents a distinct approach to modeling coreference relationships, with varying degrees of complexity and reliance on different types of information.

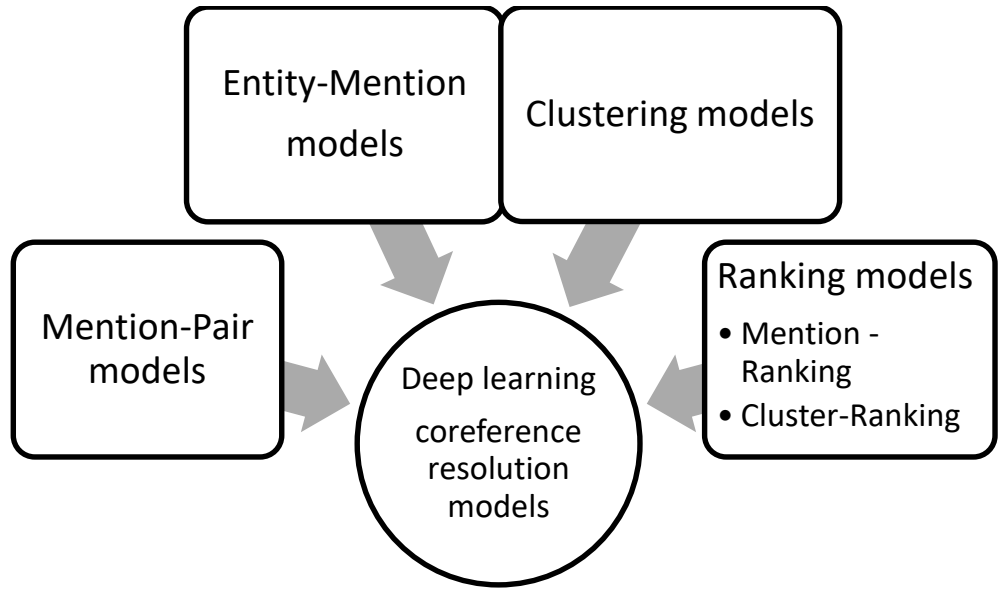

Figure 4. Deep learning coreference resolution models

### *4.3.1. Early neural network models: Pioneering Non-Linearity*

(Wiseman, Rush et al. 2015) introduced a pivotal advancement with a non-linear mention-ranking model that leveraged pre-training on referentiality identification and coreference resolution tasks. This model employed a neural network to learn an intermediate representation from raw, un-conjoined features, enabling the system to discover feature interactions automatically. This departed from the explicit feature engineering prevalent in earlier models and resulted in a significant performance improvement on the CoNLL benchmark. In subsequent work, (Wiseman, Rush et al. 2016) expanded upon this approach, introducing a non-linear cluster-based model that utilized long short-term memory (LSTM) networks to incorporate global contextual information. This enabled the model to leverage past decisions when making new ones, resulting in further performance gains.

(Clark and Manning 2016) presented another landmark contribution with their cluster-ranking model. This system treated each mention as a potential cluster, incrementally combining them based on learned representations and a single-layer neural network. The model's ability to learn and incorporate global cluster information led to state-of-the-art results on the CoNLL-2012 English dataset. As shown in Figure 5, the cluster-pair encoder uses pooling around mention-pairs to generate an entity representation.

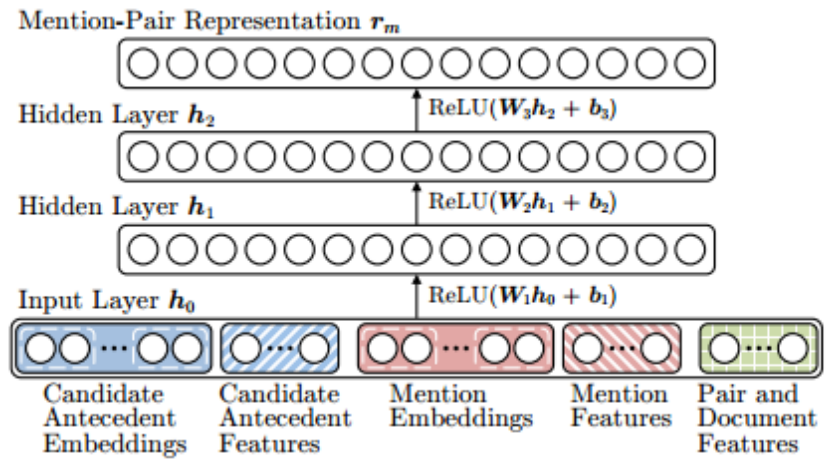

Figure 5 Mention-pair encoder derived from Clark and Manning (Clark and Manning 2016)

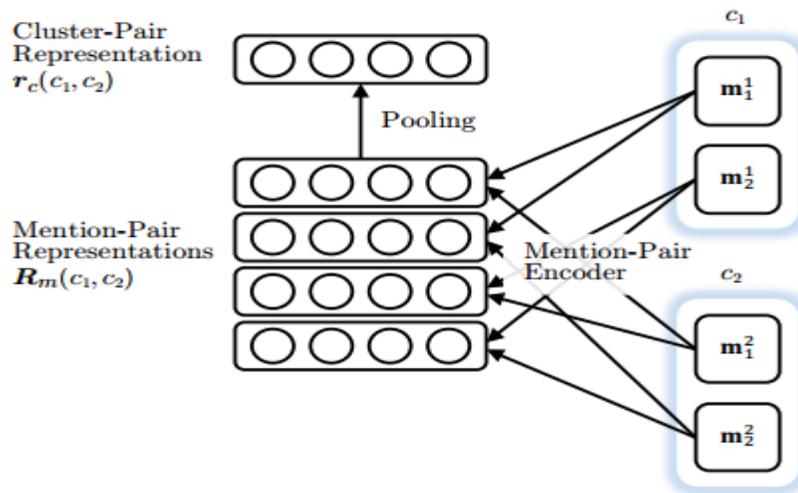

Figure 6. Cluster-pair encoder derived from Clark and Manning (Clark and Manning 2016)

Furthermore,(Clark and Manning 2016) explored using reinforcement learning (RL) to directly optimize the model's objective function based on evaluation metrics. This innovative approach showcased the potential of RL for fine-tuning neural CR models and opened up new avenues for integrating learning and evaluation into a unified framework. Their model architecture is shown in Figure 6. Here are the advantages and disadvantages of these models:

**Advantages:**

- **Non-linearity and Expressiveness:** Neural networks' ability to model non-linear relationships allows them to capture complex interactions between linguistic features, leading to improved performance compared to linear models.

- **Automatic Feature Learning:** Neural models can learn feature representations directly from raw data, reducing the reliance on manual feature engineering and potentially discovering new, informative patterns.
- **Incorporation of Global Context:** Through mechanisms like LSTMs, neural models can effectively integrate global contextual information, leading to more accurate coreference decisions.
- **Adaptability:** Neural architectures can be adapted and extended to incorporate new linguistic features, representations, or learning strategies, making them versatile tools for CR research.

**Disadvantages:**

- **Data Dependency:** Neural models typically require large amounts of annotated data for training, which can be a bottleneck for languages or domains with limited resources.
- **Interpretability:** Neural networks' internal workings are often opaque, making it difficult to understand the reasoning behind specific coreference decisions.
- **Computational Cost:** Training and inference with complex neural models can be computationally intensive, requiring specialized hardware or optimized algorithms for efficient deployment.
- **Local Optima:** The training of neural networks can get stuck in local optima, hindering the model's ability to reach its full potential performance.

*4.3.2. End-to-End Models: A Deep Dive into Neural Architectures for Coreference Resolution*

End-to-end neural models have emerged as a dominant coreference resolution (CR) paradigm, offering streamlined architectures that jointly learn mention detection and coreference linking. These models have consistently pushed the boundaries of performance, leveraging advancements in deep learning and natural language processing. We categorize and analyze prominent end-to-end CR models based on their underlying architectural principles:

**4.3.2.1 Mention-Pair Architectures: Pairwise Ranking and Classification**

Mention-pair architectures, like the seminal work of (Lee, He et al. 2017), treat CR as a ranking problem. Each mention is compared against all potential antecedents, and a score is assigned to each mention-antecedent pair using a scoring model. The antecedent with the highest score is then selected. While computationally efficient, this approach needs to work on global inference and can lead to inconsistent clustering due to its local decision-making nature. Figure shows the architecture of the end-to-end system and Figure 8 shows the Antecedent scoring of this system.

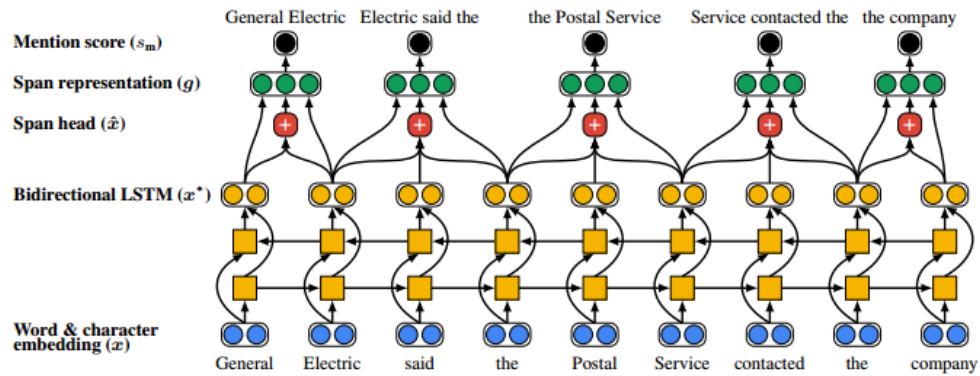

**Figure 7.** End-to-end model architecture derived from Lee et al. (Lee, He et al. 2017)

**Key Advantages:**

- **Efficiency:** Mention-pair models can be computationally efficient, especially with techniques like candidate pruning (e.g., only considering antecedents within a certain distance) and efficient scoring mechanisms.
- **Flexibility:** These models can easily incorporate various features and representations, including contextualized word embeddings (e.g., from BERT or ELMo) and attention mechanisms, to capture rich linguistic information.
- **Strong Performance:** Mention-ranking models have consistently achieved state-of-the-art results on benchmark datasets, demonstrating their effectiveness in modeling pairwise coreference relationships.

**Recent Advancements:**

- **End-to-End Neural Coreference Resolution Revisited: A Simple Yet Effective Baseline (Lee, He et al. 2017)**: This work revisits the basic mention-pair architecture and achieves competitive performance by carefully designing a model with minimal architectural complexity.
- **Coreference Resolution Through a Seq2Seq Transition-Based System** (Bohnet, Alberti et al. 2023): This approach uses a transition-based system that generates coreference chains incrementally. It achieves solid results on standard benchmarks while operating in a sequence-to-sequence fashion.
- **Higher-Order Coreference Resolution with Coarse-to-Fine Inference (Lee, He et al. 2018)**: This seminal work addresses the limitations of first-order mention-pair models by introducing a fully differentiable approximation to higher-order inference. The model uses the antecedent distribution from a span-ranking architecture as an attention mechanism to iteratively refine span representations iteratively, enabling it to capture more complex coreference relationships. A coarse-to-fine approach is employed to improve computational efficiency without sacrificing accuracy. This model significantly improved performance on the English OntoNotes benchmark and has been highly influential in subsequent CR research.

**Future Directions:**

The mention-pair paradigm for CR is still a fertile ground for research, offering ample opportunities for further exploration and improvement. Potential research directions include:

- **Exploring Alternative Architectures:** While transformer-based models have shown promising results, other mention-pair architectures, such as those based on recurrent or convolutional neural networks, could be investigated for their potential benefits.
- **Incorporating External Knowledge:** Integrating external knowledge sources, such as knowledge graphs or ontologies, could enhance the model's ability to resolve ambiguous references and capture domain-specific coreference patterns.
- **Improving Decoding Strategies:** Investigating different decoding strategies, such as beam search or reinforcement learning-based methods, could lead to more accurate and coherent coreference chain generation.
- **Standardized Evaluation and Comparison:** As highlighted by (Porada, Zou et al. 2024), the lack of a standardized experimental setup makes it difficult to compare the performance of different CR models directly. Future research should prioritize developing standardized benchmarks and evaluation protocols to facilitate more rigorous and meaningful comparisons across various mention-pair architectures.

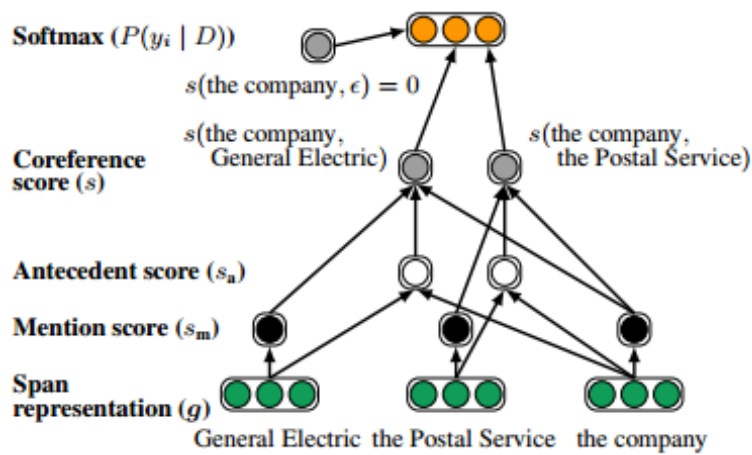

**Figure 8.** Antecedent scoring derived from Lee et al. (Lee, He et al. 2017)

### 4.3.2.2 Cluster-Ranking Architectures: Balancing Local and Global Information

Cluster-ranking models address the limitations of pure mention-pair models by introducing a two-stage approach to coreference resolution. First, mentions are clustered based on simple heuristics or pairwise similarities. Then, each cluster's ranking model is applied to determine the most likely antecedent for each mention. This approach combines the computational efficiency of mention-pair models with the ability to incorporate global information at the cluster level.

**Key Advantages:**

- **Global Coherence:** Cluster-ranking models can enforce global coherence in coreference chains by considering clusters of mentions rather than individual pairs. This ensures that coreference decisions are consistent across the document, leading to more accurate and interpretable results.
- **Rich Contextualization:** Cluster-ranking models can leverage more contextual information than mention-pair models. This includes local features (e.g., mention similarity, distance) and global features (e.g., cluster size, entity-level features).
- **Reduced Cascading Errors:** The clustering step helps to mitigate the cascading error problem, where incorrect pairwise decisions can propagate and lead to further errors in coreference chains.
- **Flexible Architecture:** Cluster-ranking models can incorporate various ranking models, including deep learning architectures based on transformers or graph neural networks, allowing for customization and flexibility.

**Recent Advancements:**

- **Anisotropic Span Embeddings and the Negative Impact of Higher-Order Inference for Coreference Resolution: An Empirical Analysis(Hou, Wang et al. 2024)**: This paper provides a comprehensive analysis of the impact of higher-order inference in cluster-ranking models, suggesting that anisotropic span embeddings can mitigate the adverse effects and improve performance.
- **SpanBERT for Coreference Resolution (Joshi, Chen et al. 2020)**: This work adapts the BERT model for span-level representations and applies it to a cluster-ranking model, achieving good results.
- **Coreference Resolution Based on High-Dimensional Multi-Scale Information** (Wang, Ding et al. 2024): This study leverages a multi-scale context information module within a BERT-based architecture to enhance the model's ability to capture global information under different text spans, demonstrating the potential of utilizing diverse contextual information in cluster-ranking models.
- **CR-M-SpanBERT: Multiple Embedding-Based DNN Coreference Resolution Using Self-Attention SpanBERT** (Jung 2024): This study introduces a new approach to cluster-ranking that utilizes multiple embeddings and the SpanBERT model. It focuses on antecedent recognition in natural language text, particularly in the context of information extraction. By generating multiple embeddings that incorporate syntactic and semantic information, this model aims to improve the accuracy of identifying mentions referring to the same entity. The study reports that CR-M-SpanBERT achieves higher recognition accuracy with fewer training epochs compared to traditional SpanBERT-based approaches.

**Future Directions in Cluster-Ranking:**

- **Dynamic Clustering:** Explore dynamic clustering methods that adapt the cluster structure as the model processes the text, potentially improving the handling of complex coreference relationships and reducing cascading errors.
- **Incorporating Discourse Structure:** Integrate discourse structure information into the clustering process to enhance the model's understanding of the underlying relationships between mentions and improve the coherence of coreference chains.
- **Leveraging Knowledge Graphs:** Utilize external knowledge graphs to augment the model's understanding of entities and their relationships, improving its ability to resolve ambiguous references and generalize to new domains.
- **Multi-Task Learning:** Combine coreference resolution with other NLP tasks, such as relation extraction or event detection, to leverage shared knowledge and improve overall performance.

### 4.3.2.3 Sequence-to-Sequence (Seq2Seq) Architectures: Generation-Based CR

Sequence-to-Sequence (Seq2Seq) models offer a fundamentally different paradigm for coreference resolution (CR), framing it as a text generation task rather than a classification or ranking problem. The input is the raw text, and the model learns to generate a modified version of the text where coreferent mentions are replaced with their canonical forms or cluster IDs. This approach simplifies the CR pipeline, eliminating the need for separate mention detection and coreference linking stages.

**Key Advantages:**

- **Flexible Handling of Coreference Phenomena:** Seq2Seq models can naturally handle diverse coreference phenomena, including nested mentions, non-contiguous entities, and bridging anaphora. Their generative nature allows for greater flexibility in representing complex coreference relationships.
- **End-to-End Training:** Seq2Seq models can be trained end-to-end, directly mapping the input text to the output coreference representation. This eliminates the need for handcrafted features or complex intermediate representations, potentially simplifying the modeling process.
- **Potential for Multilingual CR:** The Seq2Seq framework can be readily adapted to different languages, as it learns to generate coreference chains directly from text, without relying on language-specific features or rules.

Early explorations of Seq2Seq for CR, such as the work by (Urbizu, Soraluze et al. 2020), demonstrated the feasibility of this approach and its potential for handling complex coreference structures. (Fu, Song et al. 2021) extended the Seq2Seq framework to Abstract Meaning Representation (AMR) coreference resolution, showcasing its adaptability to structured semantic representations.

**Recent Advancements:**

- **Seq2seq Is All You Need for Coreference Resolution(Zhang, Wiseman et al. 2023)**: This paper establishes a strong baseline for Seq2Seq CR, demonstrating that a carefully designed Seq2Seq model with minimal architectural complexity can achieve state-of-the-art performance on multiple CR datasets.

**Future Directions:**

The Seq2Seq paradigm for CR is still relatively new, offering ample opportunities for further exploration and improvement. Potential research directions include:

- **Exploring Alternative Architectures:** While transformer-based models have shown promising results, other Seq2Seq architectures, such as those based on recurrent or convolutional neural networks, could be investigated for their potential benefits.
- **Incorporating External Knowledge:** Integrating external knowledge sources, such as knowledge graphs or ontologies, could enhance the model's ability to resolve ambiguous references and capture domain-specific coreference patterns.
- **Improving Decoding Strategies:** Investigating different decoding strategies, such as beam search or reinforcement learning-based methods, could lead to more accurate and coherent coreference chain generation.

### 4.3.2.4 Graph Neural Network (GNN) Architectures: Exploiting Structural Dependencies

Graph Neural Network (GNN) architectures have emerged as a promising direction in end-to-end coreference resolution. They leverage the inherent graph structure of coreference relationships to capture complex dependencies and global context. In these models, mentions are represented as nodes and potential coreference links are represented as edges. GNNs propagate information along these edges, allowing nodes to aggregate information from their neighbors and learn representations that reflect the broader discourse context.

**Key Advantages:**

- **Explicitly Model Relationships:** Unlike other end-to-end models, GNNs explicitly model the pairwise relationships between mentions as edges in a graph. This allows for a more direct representation of the coreference structure, enabling the model to learn more nuanced patterns of coreference.
- **Enhanced Global Context Awareness:** GNNs capture global context by propagating information across the graph. This enables each mention's representation to be enriched with information from all other mentions in the document, leading to more informed coreference decisions.
- **Flexible Representation Learning:** GNNs can learn rich representations of mentions and entities that incorporate local features (e.g., mention attributes) and global context (e.g., discourse structure). This allows a more comprehensive understanding of the semantic and syntactic relationships between mentions.

- **Potential for Incorporating External Knowledge:** GNNs can be easily extended to incorporate external knowledge sources, such as knowledge graphs or ontologies. This can provide additional information about entities and their relationships, improving coreference resolution accuracy.

**Recent Advancements:**

- **A GNN-based Coreference Resolution System (Liu, Song et al. 2020)**: This work explores using graph neural networks for CR, achieving promising results on the CoNLL-2012 dataset. It demonstrates the effectiveness of GNNs in capturing the structural dependencies of coreference chains.
- **Higher-order Inference for Coreference Resolution (Xu and Choi 2020)**: This paper investigates different higher-order inference methods for GNN-based CR, further improving performance on benchmark datasets. It highlights the importance of considering higher-order relationships between mentions for accurate coreference resolution.
- **Graph Refinement for Improved Coreference Resolution:** Exploring graph refinement techniques, such as those proposed by (Miculicich and Henderson 2022), which iteratively refine a graph-based representation of coreference links to capture document-level dependencies, could lead to significant improvements in model accuracy and robustness.
- **Heterogeneous Graph Attention Networks** (Jiang and Cohn 2021): This paper introduces a heterogeneous graph-based model (HGAT) that incorporates syntactic and semantic information to model different mentions and their relationships, improving performance on complex coreference scenarios.

**Future Directions:**

Applying GNNs to CR is a rapidly evolving area with immense potential for further advancement. Future research directions include:

- **Incorporating More Diverse Graph Structures:** Exploring different graph topologies beyond simple undirected graphs (e.g., directed graphs, hypergraphs) and incorporating additional types of nodes and edges (e.g., representing discourse relations or semantic roles) could help capture a broader range of linguistic phenomena and coreference patterns.
- **Integrating Knowledge-Aware GNNs:** Leveraging external knowledge sources, such as knowledge graphs or ontologies, could further enhance the model's understanding of entities and their relationships. This could be particularly beneficial for resolving ambiguous references and handling domain-specific coreference.
- **Developing Explainable GNNs:** A significant challenge in GNN-based CR is more interpretability. Future research should focus on developing techniques to explain the reasoning behind GNN-based coreference decisions, making the models more transparent and trustworthy.
- **Graph Refinement and Dynamic Graphs:** Exploring graph refinement techniques, such as those proposed by (Miculicich and Henderson 2022), which iteratively refine a graph-based representation of coreference links to capture document-level dependencies, could lead to significant improvements in model accuracy and robustness. Additionally, investigating dynamic graph structures that evolve as the model processes the text could further enhance the representation of coreference relationships.

## 4.4. Coreference resolution results

Table 4 compares the performance of various deep learning coreference resolution (CR) systems across different evaluation metrics and datasets. The systems are categorized into early neural networks and end-to-end models, reflecting the evolution of architectures and techniques in this field.

The first generation of neural CR systems, such as those proposed by (Wiseman, Rush et al. 2015) and (Clark and Manning 2016), demonstrated the potential of non-linear models and learned representations for improving coreference resolution. While often relying on hand-crafted features and pipeline architectures, these early models paved the way for developing more sophisticated end-to-end systems. End-to-end models, pioneered by (Lee, He et al. 2017), have emerged as the dominant paradigm in recent years, consistently achieving state-of-the-art results on benchmark datasets like CoNLL-2012. These models streamline the CR pipeline by jointly learning mention detection and coreference linking, improving efficiency and accuracy.

The evolution of end-to-end models has seen several key advancements:

- **Higher-Order Inference:** Models like (Lee, He et al. 2018) incorporate higher-order inference mechanisms to capture more complex dependencies between mentions, significantly improving performance.
- **Global Contextualization:** Techniques like attention mechanisms and graph neural networks have been leveraged to incorporate global context into decision-making, leading to more accurate and coherent coreference chains.
- **Pretrained Language Models:** systems have successfully utilized pre-trained language models like BERT and SpanBERT (Joshi, Chen et al. 2020) to enhance mention representations and improve CR performance.

As Table 4 illustrates, the performance of CR systems has steadily improved over time, with end-to-end models consistently outperforming earlier neural and rule-based approaches. Higher-order inference, global contextualization, and pre-trained language models have been instrumental in achieving these gains.

However, there are still several challenges and areas for further exploration:

- **Model Complexity:** Many state-of-the-art models are computationally expensive and require substantial resources for training and inference. This raises questions about their scalability and applicability to real-world scenarios with limited resources.
- **Evaluation Metrics:** The choice of evaluation metric can significantly impact system performance assessment. Recent research (Porada, Zou et al. 2024) highlighted that the lack of a standardized experimental setup and the limitations of existing metrics necessitate a more nuanced and comprehensive approach to evaluation.

- **Interpretability:** Neural models often lack transparency, making understanding the reasoning behind their decisions difficult. Developing more interpretable models is crucial for building trust and ensuring fairness in CR applications.

The field of CR is rapidly evolving, with ongoing research exploring new architectures, learning paradigms, and evaluation methodologies. Some promising future directions include:

- **Lightweight and Efficient Models:** Developing more efficient and lightweight models that can achieve competitive performance with fewer parameters and computational resources.
- **Hybrid Approaches:** Combining the strengths of different architectures, such as integrating mention-pair and cluster-ranking models, could lead to further improvements.
- **Incorporating External Knowledge:** Leveraging external knowledge sources, such as knowledge graphs or ontologies, could help resolve ambiguous references and enhance the model's understanding of entity relationships.
- **Task-Specific Fine-tuning:** Adapting CR models to specific domains and tasks (e.g., biomedical text, social media) can improve their performance and applicability in real-world scenarios.

By addressing these challenges and pursuing these promising directions, researchers can continue to push the boundaries of coreference resolution, leading to systems that more accurately and reliably understand the complex relationships between entities and events in natural language text.

**Table 4.** Results of coreference resolution systems

| Deep learning system | MUC | B3 | CEAF | CoNLL | Implementation |
|---|---|---|---|---|---|
| (Fernandes, Dos Santos et al. 2012) | 70.51 | 57.58 | 53.86 | 60.65 | N/A |
| (Durrett and Klein 2013) | 70.51 | 58.33 | 55.36 | 61.4 | N/A |
| (Björkelund and Kuhn 2014) | 70.72 | 58.58 | 55.61 | 61.63 | N/A |
| (Durrett and Klein 2014) | 71.24 | 58.71 | 55.18 | 61.71 | N/A |
| (Wiseman, Rush et al. 2015) | 72.0 | 60.5 | 57.1 | 63.4 | https://github.com/swiseman/nn_coref |
| (Wiseman, Rush et al. 2016) | 73.4 | 61.5 | 57.7 | 64.2 | https://github.com/swiseman/nn_coref |
| (Clark and Manning 2016) | 74.0 | 62.9 | 59.0 | 65.3 | https://github.com/clarkkev/deep-coref |
| (Clark and Manning 2016) | 74.6 | 63.4 | 59.2 | 65.7 | https://github.com/clarkkev/deep-coref |
| (Meng and Rumshisky 2018) | 80.9 | 65.7 | 50.8 | 65.8 | https://github.com/text-machine-lab/entity-coref |
| (Swayamdipta, Thomson et al. 2018) | 76.3 | 65.7 | 61.5 | 67.8 | https://github.com/swabhs/scaffolding |
| (Luo and Glass 2018) | 76.4 | 65.6 | 61.4 | 67.8 | https://github.com/luohongyin/coatt-coref |
| (Lee, He et al. 2017) | 77.2 | 66.6 | 62.6 | 68.8 | https://github.com/kentonl/e2e-coref |
| (Zhang, Santos et al. 2018) | 77.6 | 67.1 | 62.9 | 69.2 | N/A |
| (Lee, He et al. 2018) | 80.4 | 70.8 | 67.6 | 73.0 | https://github.com/kentonl/e2e-coref |
| (Subramanian and Roth 2019) | 80.7 | 71.1 | 67.9 | 73.2 | N/A |
| (Fei, Li et al. 2019) | 81.4 | 71.7 | 68.4 | 73.8 | N/A |
| (Kantor and Globerson 2019) | 83.4 | 74.7 | 71.8 | 76.6 | https://github.com/bkntr/coref-ee |
| (Joshi, Levy et al. 2019) | 83.5 | 75.3 | 71.9 | 76.9 | https://github.com/mandarjoshi90/coref |
| (Liu, Song et al. 2020) | 83.8 | 75.1 | 72.2 | 77 | N/A |
| (Toshniwal, Wiseman et al. 2020) | 84.7 | 76.8 | 73.2 | 78.2 | https://github.com/shtoshni92/long-doc-coref |
| (Xia, Sedoc et al. 2020) | 85.3 | 77.8 | 75.2 | 79.4 | https://github.com/pitrack/incremental-coref |
| (Joshi, Chen et al. 2020) | 85.3 | 78.1 | 75.3 | 79.6 | https://github.com/facebookresearch/SpanBERT |
| (Lai, Bui et al. 2022) | 85.4 | 78.7 | 75.0 | 79.7 | N/A |
| (Wang, Ding et al. 2024) | 85.4 | 78.4 | 76.1 | 79.8 | N/A |
| (Hourali, Zahedi et al. 2020) | 86.1 | 79.2 | 74.8 | 80 | N/A |
| (Xu and Choi 2020) | 85.7 | 79 | 75.9 | 80.2 | https://github.com/lxucs/coref-hoi |
| (Jung 2024) | 85.5 | 79.5 | 76.2 | 80.4 | N/A |
| (Miculicich and Henderson 2022) | 85.9 | 79.3 | 76.4 | 80.5 | N/A |
| (Jiang and Cohn 2021) | 86.5 | 78 | 76.5 | 81.1 | https://github.com/Fantabulous-J/coref-HGAT |
| (Hou, Wang et al. 2024) | 86.3 | 80.7 | 77.9 | 81.6 | N/A |
| (Wu, Wang et al. 2020) | 88 | 82.2 | 79.1 | 83.1 | https://github.com/ShannonAI/CorefQA |
| (Zhang, Wiseman et al. 2023) | 87.6 | 82.4 | 79.5 | 83.2 | https://github.com/WenzhengZhang/Seq2seqCoref |
| (Bohnet, Alberti et al. 2023) | 87.8 | 82.6 | 79.5 | 83.3 | https://github.com/google-research/google-research/tree/master/coref_mt5 |
| (Khosla and Rose 2020) | 92.2 | 84.4 | 79.9 | 85.5 | N/A |
| (Wang, Shen et al. 2021) | 92.5 | 85.9 | 84.1 | **87.5** | N/A |

## 5. Persian coreference resolution systems

Persian, with its distinct lexical, syntactic, and morphological structure compared to English, presents unique challenges and opportunities for coreference resolution (CR). Notably, the absence of gendered pronouns in Persian necessitates different approaches compared to English-centric models.

The first attempts at Persian CR focused on pronoun resolution, with )Moosavi and Ghassem-Sani 2009) evaluating various machine learning algorithms on a manually annotated corpus derived from the Bijankhan corpus. While limited in performance, this work laid the groundwork for further exploration. Subsequent studies (Nourbakhsh and Bahrani 2017) refined feature sets and improved results, while introduced the first rule-based Persian CR system, achieving high precision and recall on a small corpus of blog data.

The development of the RCDAT corpus (Rahimi and HosseinNejad 2020) , a large-scale annotated dataset containing approximately one million tokens, reached a significant milestone. This corpus, adhering to the CoNLL standard, facilitated developing and evaluating more sophisticated Persian CR systems

The availability of the RCDAT corpus spurred the development of several Persian CR systems. (Rahimi and HosseinNejad 2020) ) presented a baseline system using a support vector machine, achieving an average CoNLL score of 60 on gold standard annotations. (Haji Mohammadi, Talebpour et al. 2023) introduced the Mehr corpus, a meticulously annotated Persian dataset that addresses some shortcomings of previous corpora and boasts a high inter-annotator agreement. This corpus annotates coreference relations for various linguistic elements, including noun phrases, named entities, pronouns, and nested named entities. (Sahlani, Hourali et al. 2020, Sahlani, Hourali et al. 2020) leveraged a fully connected deep neural network for feature extraction and a hierarchical clustering algorithm for coreference chain generation, improving upon previous results on the Uppsala Persian test dataset (Seraji, Megyesi et al. 2012).

The emergence of end-to-end neural models (Lee, He et al. 2017) has opened up new avenues for Persian CR. These models eliminate the need for handcrafted features and syntactic parsers, offering a streamlined approach to jointly learning mention detection and coreference linking. The current study by (Mohammadi, Talebpour et al. 2024) presents the first end-to-end neural system for Persian pronoun resolution, leveraging pre-trained Transformer models like ParsBERT (Farahani, Gharachorloo et al. 2021). By jointly optimizing mention detection and antecedent linking, their system achieves a substantial improvement over previous state-of-the-art systems on the Mehr corpus.

Despite the progress made in Persian CR, several challenges and opportunities remain:

- **Data Scarcity:** The availability of large-scale, high-quality annotated corpora for Persian is still limited compared to English. This hinders the development and evaluation of more sophisticated models.
- **Linguistic Complexity:** Persian exhibits unique linguistic features that pose challenges for CR:
    - **Complex Verb Morphology:** Persian verbs inflect for tense, aspect, mood, person, and number, creating many surface forms. This morphological complexity makes it challenging to identify and match verb mentions for coreference resolution accurately.
    - Pro-Drop: Persian is a pro-drop language, meaning pronouns can be omitted when their referents are understood from context. This introduces ambiguity and increases the difficulty of identifying coreference relationships.
- **Word Embedding Challenges:**
    - **Limited Resources:** While pre-trained word embeddings exist for Persian (e.g., ParsBERT), their quality and coverage may differ from those available for English. This can limit their effectiveness in capturing semantic and contextual information necessary for accurate coreference resolution.
    - **Morphological Complexity:** The rich morphology of Persian can lead to challenges in creating effective word embeddings. Inflectional variations and compound words may need to be better represented in standard word embedding models, impacting the ability to capture semantic relationships between mentions.
    - **Domain Adaptation:** Pre-trained word embeddings are often trained on general-domain text, which may not adequately capture specialized domains' specific vocabulary and usage patterns. Adapting these embeddings to specific domains or tasks can be crucial for improving CR performance in those contexts.
- **Evaluation and Benchmarking:** Standardized evaluation benchmarks and datasets are needed for Persian CR to facilitate fair comparison and progress in this area.

Future research should focus on:

- **Expanding Annotated Resources:** Creating more extensive and diverse corpora that better reflect Persian's rich linguistic landscape. This includes exploring new annotation schemes, incorporating various genres and domains, and addressing biases in existing datasets.
- **Incorporating External Knowledge:** Leveraging resources like Persian word embeddings, knowledge graphs, and ontologies can enhance the model's understanding of entities and their relationships. This can be particularly helpful in disambiguating references and resolving coreference across different sentences or documents.
- **Tailoring Models to Persian:** Developing CR models addressing Persian-specific linguistic challenges, such as pro-drop and complex verb morphology. This may involve incorporating morphological features, developing specialized syntactic parsers, or adapting existing models to handle these nuances better.
- **Leveraging Advanced NLP Techniques:** Exploring the application of cutting-edge NLP techniques to Persian CR, such as:
    - **Transformer-based Models:** Building upon the success of pre-trained transformer models like ParsBERT, researchers can fine-tune these models for Persian CR and investigate their effectiveness in capturing complex linguistic dependencies.
    - **Attention Mechanisms:** Incorporating attention mechanisms into CR models can help focus on relevant contextual information and improve the accuracy of coreference resolution.
    - **Sequence-to-Sequence (Seq2Seq) Models:** The Seq2Seq framework, which has shown promise in other NLP tasks, could be explored for its potential in Persian CR, mainly for handling non-contiguous entities and generating coreference chains directly from text.

By addressing these challenges and pursuing these avenues, researchers can significantly advance the field of Persian coreference resolution, contributing to the broader goal of developing effective NLP systems for underrepresented languages.

## 6. Discussion and issues

Substantial progress has been made in the evolution of coreference resolution (CR), particularly with the rise of deep learning. However, challenges persist, and new opportunities emerge in this dynamic field.

- **Feature Representation and Global Context:** While deep learning has reduced the reliance on handcrafted features, effective representation learning remains challenging. Early neural models often struggled to capture complex features (Wiseman, Rush et al. 2015). Recent advances, such as biaffine attention (Zhang, dos Santos et al. 2018), have shown promise, but further exploration is needed, especially for low-resource languages like Persian. Incorporating global context remains crucial. Higher-order inference (Lee, He et al. 2018) and graph neural networks (Liu, Song et al. 2020, Xu and Choi 2020, Miculicich and Henderson 2022) have improved the modeling of long-range dependencies, but scaling to more significant documents and handling diverse linguistic phenomena, especially in morphologically rich languages like Persian, are ongoing challenges.

- **End-to-End Models: Joint Learning and Global Inference:** End-to-end models (Lee, He et al. 2017) have simplified the CR pipeline, but maintaining global coherence and avoiding inconsistencies are ongoing challenges. Recent studies (Liu, Song et al. 2020, Bohnet, Alberti et al. 2023, Zhang, Wiseman et al. 2023) have proposed new architectures and training strategies to address these issues, showcasing the continued innovation in this area. However, as noted by (Porada, Zou et al. 2024), a standardized experimental setup must make direct comparisons between models easier, hindering progress. This highlights the need for standardized benchmarks and evaluation protocols in future research.
- **Generalization and Domain Adaptation:** Generalizing CR models across domains and languages is critical. While techniques like dropout (Lee, Surdeanu et al. 2017) and adversarial training (Subramanian and Roth 2019) have been employed, more robust solutions are needed. Transfer learning and domain adaptation hold promise, especially for low-resource languages like Persian. The work of (Demir 2023) on Turkish CR provides valuable insights into leveraging multilingual transfer learning for under-resourced languages.
- **Evaluation and Beyond:** The CoNLL metric, though widely used, has limitations (Luo 2005). Developing more comprehensive and task-specific evaluation metrics is crucial. Additionally, future research should focus on:
  - **Event Coreference:** Expanding CR to encompass event coreference, as recent work has shown promising results in this direction (Hürriyetoğlu, Mutlu et al. 2022, Lu, Lin et al. 2022, Yu, Yin et al. 2022).
  - **Document-Level Context:** We are investigating models that capture document-level context to better resolve long-range coreference dependencies and handle complex discourse structures.
  - **Interpretability and Explainability:** Developing methods to make CR models more interpretable, enabling users to understand the reasoning behind coreference decisions and build trust in the system's outputs.

By tackling these challenges and exploring these promising avenues, the field of coreference resolution will continue to evolve, leading to more accurate, robust, and interpretable systems that can effectively understand and process the complex relationships within natural language text.

## 7. Conclusion

Coreference resolution, a cornerstone of natural language understanding, has evolved remarkably from its origins in rule-based systems to the current era of sophisticated deep learning models.

This comprehensive review has traversed this journey, examining the diverse corpora that fuel CR research, the evolving landscape of evaluation metrics, and the intricate architectures of various CR models. We have highlighted the strengths and limitations of each approach, emphasizing the critical role of global context, feature representation, and end-to-end learning in pushing the boundaries of CR performance. We have also delved into the specific challenges and opportunities in Persian CR, showcasing recent advancements in this under-resourced language.

Despite significant progress, several challenges still need to be addressed, including data scarcity, linguistic complexities, model interpretability, and more nuanced evaluation metrics. These challenges present exciting opportunities for future research, with promising avenues including exploring diverse graph structures, integrating external knowledge, developing explainable models, and creating more comprehensive benchmarks. The insights from this review will serve as a valuable resource for researchers and practitioners, guiding them in developing more accurate, robust, and interpretable CR systems. By addressing the remaining challenges and embracing the emerging opportunities, we can pave the way for the next generation of coreference resolution technologies to understand and process the rich tapestry of human language truly. This, in turn, will unlock new possibilities for NLP applications in areas such as information extraction, machine translation, question answering, and beyond.

Finkel, J. R. and C. D. Manning (2008). Enforcing transitivity in coreference resolution. Proceedings of the 46th Annual Meeting of the Association for Computational Linguistics on Human Language Technologies: Short Papers, Association for Computational Linguistics.

Fu, Q., et al. (2021). End-to-end AMR coreference resolution. Proceedings of the 59th Annual Meeting of the Association for Computational Linguistics and the 11th International Joint Conference on Natural Language Processing (Volume 1: Long Papers).

Gaizauskas, R., et al. (1995). UNIVERSITY OF SHEFFIELD: DESCRIPTION OF THE LaSIE SYSTEMAS USED FOR MUC-6. Sixth Message Understanding Conference (MUC-6): Proceedings of a Conference Held in Columbia, Maryland, November 6-8, 1995.

Gao, Q., et al. (2024). Enhancing Cross-Document Event Coreference Resolution by Discourse Structure and Semantic Information. Proceedings of the 2024 Joint International Conference on Computational Linguistics, Language Resources and Evaluation (LREC-COLING 2024).

Ge, N., et al. (1998). A statistical approach to anaphora resolution. Sixth Workshop on Very Large Corpora.

Ghaddar, A. and P. Langlais (2016). Wikicoref: An english coreference-annotated corpus of wikipedia articles. Proceedings of the Tenth International Conference on Language Resources and Evaluation (LREC 2016).

Grishman, R. and B. M. Sundheim (1996). Message understanding conference-6: A brief history. COLING 1996 Volume 1: The 16th International Conference on Computational Linguistics.

Guillou, L., et al. (2014). ParCor 1.0: A parallel pronoun-coreference corpus to support statistical MT. 9th International Conference on Language Resources and Evaluation (LREC), MAY 26-31, 2014, Reykjavik, ICELAND, European Language Resources Association.

Haji Mohammadi, H., et al. (2023). "Mehr: A Persian Coreference Resolution Corpus." Journal of AI and Data Mining.

Harabagiu, S. M., et al. (2001). Text and knowledge mining for coreference resolution. Proceedings of the second meeting of the North American Chapter of the Association for Computational Linguistics on Language technologies, Association for Computational Linguistics.

Hasler, L. and C. Orasan (2009). Do coreferential arguments make event mentions coreferential. Proceedings of the 7th Discourse Anaphora and Anaphor Resolution Colloquium (DAARC 2009), Citeseer.